\documentclass[conference]{IEEEtran}
\IEEEoverridecommandlockouts
\usepackage{cite}
\usepackage{amsmath,amssymb,amsfonts}
\usepackage{graphicx}
\usepackage{textcomp}
\usepackage{xcolor}
\usepackage{hyperref}
\usepackage{algpseudocode}
\usepackage{booktabs}
\usepackage{multirow}
\usepackage{siunitx}
\usepackage{subcaption}
\usepackage{stfloats}
\usepackage{tikz}
\usepackage{url}
\usepackage{enumitem}
\usepackage{soul}
\usepackage{colortbl,bbding,arydshln}
\usepackage[flushleft]{threeparttable}

\def\BibTeX{{\rm B\kern-.05em{\sc i\kern-.025em b}\kern-.08em
    T\kern-.1667em\lower.7ex\hbox{E}\kern-.125emX}}

\graphicspath{{img/}{figures/}}

\newcommand{\runinhead}[1]{\textbf{#1.}}

\makeatletter
\newcommand{\inputIfExists}[1]{%
  \IfFileExists{#1.tex}{\input{#1}}{%
    \PackageWarning{paper}{Missing file: #1.tex}%
  }%
}
\makeatother



\begin{document}

\title{MoCrop: Training Free Motion Guided Cropping for Efficient Video Action Recognition}

\author{Binhua Huang\textsuperscript{1},
        Wendong Yao\textsuperscript{1},
        Shaowu Chen\textsuperscript{2},
        Guoxin Wang\textsuperscript{3}, 
        Qingyuan Wang\textsuperscript{3}, 
        Soumyabrata Dev\textsuperscript{1}
        \\
        \textsuperscript{1}School of Computer Science, University College Dublin \\
        \textsuperscript{2}Undergraduate School of Artificial Intelligence, Shenzhen Polytechnic University \\
        \textsuperscript{3}School of Electrical and Electronic Engineering, University College Dublin
}

\maketitle

\begin{abstract}
Standard video action recognition models often process typically resized full frames, suffering from spatial redundancy and high computational costs. To address this, we introduce MoCrop, a motion-aware adaptive cropping module designed for efficient video action recognition in the compressed domain. Leveraging Motion Vectors (MVs) naturally available in H.264 video, MoCrop localizes motion-dense regions to produce adaptive crops at inference without requiring any training or parameter updates. Our lightweight pipeline synergizes three key components: Merge \& Denoise (MD) for outlier filtering, Monte Carlo Sampling (MCS) for efficient importance sampling, and Motion Grid Search (MGS) for optimal region localization. This design allows MoCrop to serve as a versatile ``plug-and-play'' module for diverse backbones. Extensive experiments on UCF101 demonstrate that MoCrop serves as both an accelerator and an enhancer. With ResNet-50, it achieves a +3.5\% boost in Top-1 accuracy at equivalent FLOPs (Attention Setting), or a +2.4\% accuracy gain with 26.5\% fewer FLOPs (Efficiency Setting). When applied to CoViAR, it improves accuracy to 89.2\% or reduces computation by roughly 27\% (from 11.6 to 8.5 GFLOPs). Consistent gains across MobileNet-V3, EfficientNet-B1, and Swin-B confirm its strong generality and suitability for real-time deployment. Our code and models are available at \url{https://github.com/microa/MoCrop}.
\end{abstract}


\section{Introduction}

Deep learning has fundamentally transformed video action recognition, evolving from hand-crafted features to sophisticated spatiotemporal representations learned via 3D convolutions and Video Transformers \cite{simonyan2014very, tran2015learning, liu2022video, bertasius2021space}. However, this performance leap comes with substantial computational costs. A primary source of inefficiency is spatial redundancy: standard models process the entire frame—often resized to a fixed resolution (e.g., $224 \times 224$)—regardless of whether the action occurs in a localized region or spans the whole scene. While spatial attention mechanisms \cite{guo2022attention, woo2018cbam} can mitigate this by highlighting informative areas, they typically require end-to-end training, introduce additional parameters, and are coupled tightly with specific backbone architectures.

To address efficiency without sacrificing accuracy, the compressed domain offers a compelling alternative \cite{wu2018compressed, wiegand2003overview}. Video codecs like H.264 naturally encode motion information via Motion Vectors (MVs) to reduce temporal redundancy. Prior works such as CoViAR \cite{wu2018compressed} and MV2Flow \cite{hu2020mv2flow} treat these MVs as input modalities to learn features. In contrast, we hypothesize that MVs serve a more primal purpose: they are essentially "free" saliency maps that can guide the model to focus on Spatially Salient Regions (SSR) before inference begins \cite{zheng2023dynamic, shahabinejad2023video}.

Building on this insight, we introduce MoCrop, a lightweight, training-free module that leverages codec MVs for adaptive spatial cropping. Unlike learned attention modules, MoCrop operates as a pre-processor. It addresses the inherent noisiness and sparsity of raw MVs through a novel pipeline: Merge \& Denoise (MD) to filter outliers, Monte Carlo Sampling (MCS) to drastically reduce processing overhead, and Motion Grid Search (MGS) to lock onto the optimal action region.

MoCrop offers a versatile "Plug-and-Play" capability. In an Efficiency Setting, it allows models to process smaller, cropped resolutions (e.g., $192\text{px}$), significantly reducing FLOPs. In an Attention Setting, it removes background clutter for standard inputs, boosting accuracy. Crucially, MoCrop requires no retraining of the backbone model, making it immediately applicable to diverse architectures ranging from ResNet to Swin Transformers.

Our main contributions are summarized as follows:
\begin{itemize}
    \item We propose MoCrop, a training-free, parameter-free adaptive cropping module that utilizes compressed-domain Motion Vectors to guide efficient video recognition.
    \item We design a high-efficiency pipeline incorporating Monte Carlo Sampling and dual-objective grid search, enabling robust region selection with negligible computational overhead.
    \item We demonstrate that MoCrop serves as a universal accelerator and enhancer. Experiments on UCF101 show it can improve Top-1 accuracy by up to 3.5\% (ResNet-50) or reduce FLOPs by over 26\% while maintaining accuracy, generalizing across CNNs and Transformers.
\end{itemize}

\section{Related Work}

\textbf{Video Action Recognition.}
Early approaches to video action recognition relied on hand-crafted features but struggled with scalability \cite{zhang2022hybrid}. 
The emergence of deep learning marked a paradigm shift, with the Two-Stream Network \cite{simonyan2014two} leveraging pre-computed optical flow to capture motion patterns. 
However, optical flow computation remains expensive. 
C3D \cite{tran2015learning} introduced 3D convolutions to learn spatiotemporal features end-to-end, avoiding explicit optical flow but incurring high computational costs. 
More recently, video transformers such as Video Swin \cite{liu2022video} and TimeSformer \cite{bertasius2021space} achieve state-of-the-art performance by modeling long-range dependencies through self-attention, though at significant computational expense. 
These methods motivate efficient alternatives that reduce computation while maintaining accuracy.

\textbf{Compressed-Domain Video Understanding.}
Compressed video formats like H.264 encode videos using I-frames (independently coded), P-frames (predicted from previous frames), and B-frames (bi-directionally predicted), with MVs and residuals encoding inter-frame differences \cite{wiegand2003overview}. 
Leveraging these codec-provided features offers computational advantages over processing raw pixels.
CoViAR \cite{wu2018compressed} pioneered compressed-domain action recognition by processing I-frames, MVs, and residuals in separate streams. 
MV2Flow \cite{hu2020mv2flow} refines MVs to approximate optical flow, bridging efficiency and accuracy. 
Recent works \cite{ming2024action, shahabinejad2023video} further exploit MVs to guide temporal frame selection and spatial attention, demonstrating that compressed-domain cues can enhance both efficiency and recognition performance.

\textbf{Spatial Attention and Adaptive Cropping.}
Spatial attention mechanisms identify and emphasize informative regions within frames. 
Early convolutional approaches like CBAM \cite{woo2018cbam} and SE-Net \cite{hu2018squeeze} apply channel and spatial attention modules trained end-to-end. 
Transformer-based methods \cite{bertasius2021space, dosovitskiy2020image} use self-attention to model spatial dependencies but require substantial computation and parameter overhead.
In the context of video understanding, adaptive cropping has been explored to focus on action-relevant regions. 
Traditional methods include center cropping, random cropping during training, and saliency-based ROI extraction \cite{guo2022attention}. 
Recent work by Shahabinejad et al. \cite{shahabinejad2023video} integrates MVs into learned spatial attention for compressed video, achieving improved efficiency.
However, most attention and cropping methods require training or fine-tuning for each backbone and dataset. 

\begin{figure*}[!t]
  \centering
  \includegraphics[width=\linewidth]{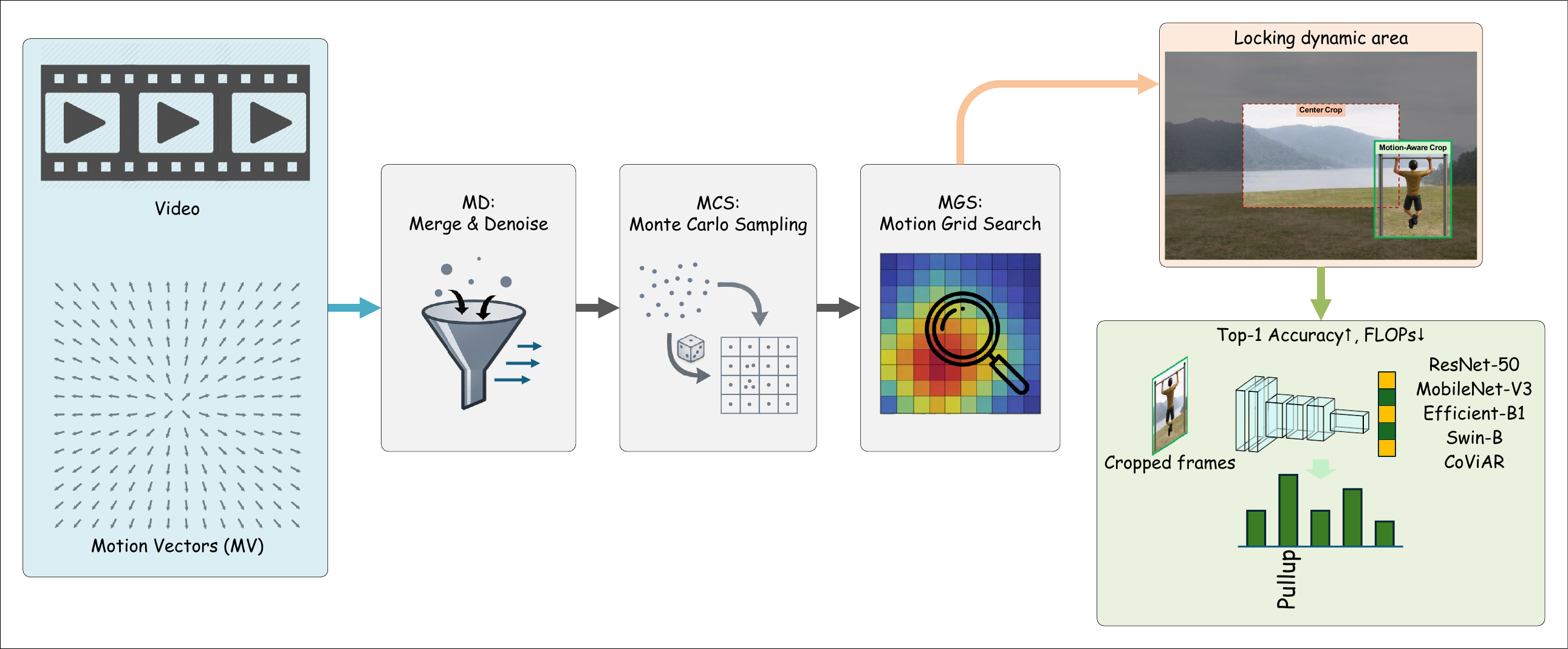}
  \caption{MoCrop pipeline. MVs identify SSR to guide I-frame cropping; cropped frames are fed to the action recognition model.}
  \label{fig:pipeline}
\end{figure*}

\textbf{Our Approach.}
Unlike prior attention mechanisms that are learned end-to-end or require retraining, MoCrop is a training-free, plug-and-play module that operates solely at inference time. 
By aggregating MVs into a motion-density map and performing a lightweight spatial search, MoCrop adaptively crops to motion-salient regions without modifying model parameters or training protocols. 
This design makes it compatible with diverse backbones and datasets, offering a practical path to efficiency and accuracy improvements in compressed-domain video action recognition.

\section{Method}

MoCrop is a preprocessing module that uses MVs from compressed streams to locate SSR and compute adaptive crops, focusing inference on informative regions. This reduces input resolution and FLOPs and often improves accuracy by suppressing background distraction.

\subsection{Overview}

Fig.~\ref{fig:pipeline} illustrates the complete MoCrop pipeline. MoCrop comprises three processing modules: \textbf{Merge \& Denoise (MD)}, \textbf{Monte Carlo Sampling (MCS)}, and \textbf{Motion Grid Search (MGS)}. MD retains top-$k$ MVs by magnitude; MCS applies weighted importance sampling to downsample filtered MVs and builds a motion-density map; MGS searches for a high-score rectangle via dual-objective optimization to determine the cropping region.

\subsection{Problem Formulation}

Given decoded I-frames and raw MVs, let $V_{\rm raw}=\{(p_k,\Delta p_k)\}_{k=1}^{T}$ with $p_k=(x_k,y_k)$ the origin and $\Delta p_k$ the displacement. We seek a function $f$ that maps MVs to a bounding box $B^\ast$:
\[
B^\ast=f(V_{\rm raw}),
\]
and each I-frame $I_i$ is then cropped by
\[
I_i'=g(B^\ast,I_i).
\]

\subsection{Detailed Algorithm}
We now describe each stage in detail.

\subsubsection{Stage 1: Merge \& Denoise (MD)}
\runinhead{Extraction and Filtering}
Let $V_{\rm raw}=\{(p_k,\Delta p_k)\}_{k=1}^{T}$ denote raw MVs across frames. We retain only the top-$k$ vectors by motion magnitude:
\[
V_{\rm filtered} = \text{top-}k(V_{\rm raw}, \alpha), \quad k = \lfloor \alpha |V_{\rm raw}| \rfloor,
\]
where $\alpha\in(0,1]$ controls the denoising strength (e.g., $\alpha{=}0.01$ keeps the top 1\%).

\subsubsection{Stage 2: Monte Carlo Sampling (MCS)}
For efficiency when $|V_{\text{filtered}}|$ is large, we apply weighted importance sampling. Each vector is sampled with probability proportional to its motion magnitude raised to power $\beta$:
\begin{align*}
p_i &\propto \|\Delta p_i\|_2^\beta, \\
V_{\text{sampled}} &\sim \text{Multinomial}(V_{\text{filtered}}, N, \{p_i\}),
\end{align*}
where $N{=}\lfloor \gamma |V_{\text{filtered}}| \rfloor$, $\gamma{\in}(0,1]$ is the sampling ratio (e.g., $\gamma{=}0.1$ samples 10\%), and $\beta{>}0$ (e.g., $\beta{=}4.0$) biases toward high-magnitude motion.

\subsubsection{Stage 3: Motion Grid Search (MGS)}

\runinhead{Grid Quantization}
We discretize the frame into an $h \times w$ grid and map each MV to its corresponding cell:
\[
M_{ij} = \sum_{(p_k, \cdot) \in V_{\text{sampled}}} \mathbb{I}(p_k \in C_{ij}),
\]
where $C_{ij}$ spans $[j \frac{W}{w}, (j+1)\frac{W}{w}) \times [i \frac{H}{h}, (i+1)\frac{H}{h})$. This motion-density map $M$ guides the subsequent search.

\runinhead{Submatrix Search}
We identify the grid region $R$ with maximum weighted score:
\begin{align*}
R^* &= \arg\max_{R \in \mathcal{R}_\rho} S(R), \\
S(R) &= w_{\text{sum}} \cdot \sum_{(i,j) \in R} M_{ij} + w_{\text{avg}} \cdot \frac{1}{|R|} \sum_{(i,j) \in R} M_{ij},
\end{align*}
where $\mathcal{R}_\rho {=} \{R : (1{-}\delta)\rho hw {\le} |R| {\le} (1{+}\delta)\rho hw\}$ constrains candidate rectangles to target area ratio $\rho$ with tolerance $\delta$, and $(w_{\text{sum}}, w_{\text{avg}})$ balance total vs. average density (e.g., $(0.6, 0.4)$). The optimal region $R^*$ is converted to normalized box coordinates for cropping.

\subsubsection{Stage 4: Cropping I-frames}
We map the grid region $R^*$ to pixel coordinates and extract the corresponding crop:
\begin{align*}
I_{\rm crop} &= I[y_1:y_2,\ x_1:x_2], \\
(x_1,y_1,x_2,y_2) &= \text{Grid2Pixel}(R^*, W_{\rm video}, H_{\rm video}).
\end{align*}
The same box is applied to all I-frames in the clip, yielding an efficient, training-free, plug-and-play preprocessor.

\subsection{Complexity and Practicality}

MD performs top-$k$ selection via \texttt{argpartition} in $O(|V_{\rm raw}|)$ time. MCS computes magnitude weights and samples in $O(|V_{\rm filtered}|)$. Building the motion-density map costs $O(|V_{\rm sampled}|{+}hw)$ for counting on an $h{\times}w$ grid. MGS enumerates all rectangles within the area constraint: with integral images, scoring each candidate is $O(1)$, and the total search is $O(h^2w^2)$ but negligible for small grids (e.g., $16{\times}9$). Overall preprocessing overhead is negligible, making MoCrop practical for real-time deployment.

\noindent\textbf{Hyperparameters.}
Key parameters include: $\alpha$ (denoising ratio, e.g., 0.01) controls noise filtering strength; $\gamma$ (sampling ratio, e.g., 0.1) trades variance for speed; $\beta$ (importance weight, e.g., 4.0) biases sampling toward high-motion regions; $(h,w)$ (grid resolution, e.g., $16{\times}9$) balances localization precision and search cost; $\rho$ (target area ratio) sets crop size; $(w_{\text{sum}}, w_{\text{avg}})$ (e.g., 0.6, 0.4) balance region compactness vs. density.

\section{Experiments}

\subsection{Dataset and Metrics}
All experiments are conducted on UCF101, a widely used benchmark for human action recognition featuring 13,320 videos across 101 action classes \cite{soomro2012ucf101}. Given that our approach is designed for efficiency, we evaluate performance by jointly considering Top-1 accuracy and computational cost in GFLOPs.

\begin{table*}[!t]
\centering
\caption{Demonstration that MoCrop provides dual advantages as a plug-and-play module on modern backbones using the UCF101 dataset. The table highlights the performance gains in both the accuracy-focused \textbf{Attention Setting} and the cost-saving \textbf{Efficiency Setting}.}
\label{tab:main_results_corrected}
\resizebox{\textwidth}{!}{%
\begin{tabular}{@{}lllcccc@{}}
\toprule
\textbf{Backbone} & \textbf{Dataset} & \textbf{Variant} & \textbf{Top-1 Acc. (\%)} & \textbf{GFLOPs} & \textbf{$\Delta$ Acc. (\%)} & \textbf{$\Delta$ FLOPs (\%)} \\
\midrule
\multirow{3}{*}{ResNet-50~\cite{he2016deep}} & \multirow{3}{*}{UCF101} & (A) Baseline (224px) & 80.1 & 4.11 & - & - \\
 & & (B) + MoCrop (Efficiency, 192px) & 82.5 & \textbf{3.02} & +2.4 & \textbf{-26.5} \\
 & & (C) + MoCrop (Attention, 224px) & \textbf{83.6} & 4.11 & \textbf{+3.5} & 0.0 \\
\midrule
\multirow{3}{*}{MobileNet-V3~\cite{howard2019searching}} & \multirow{3}{*}{UCF101} & (A) Baseline (224px) & 78.3 & 0.22 & - & - \\
 & & (B) + MoCrop (Efficiency, 192px) & 79.3 & \textbf{0.17} & +1.0 & \textbf{-22.7} \\
 & & (C) + MoCrop (Attention, 224px) & \textbf{80.2} & 0.22 & \textbf{+1.9} & 0.0 \\
\midrule
\multirow{3}{*}{EfficientNet-B1~\cite{tan2019efficientnet}} & \multirow{3}{*}{UCF101} & (A) Baseline (224px) & 82.1 & 0.59 & - & - \\
 & & (B) + MoCrop (Efficiency, 192px) & 83.6 & \textbf{0.43} & +1.5 & \textbf{-27.1} \\
 & & (C) + MoCrop (Attention, 224px) & \textbf{84.5} & 0.59 & \textbf{+2.4} & 0.0 \\
\midrule
\multirow{3}{*}{Swin-B~\cite{liu2022video}} & \multirow{3}{*}{UCF101} & (A) Baseline (224px) & 87.3 & 15.5 & - & - \\
 & & (B) + MoCrop (Efficiency, 192px) & 87.2 & \textbf{12.6} & -0.1 & \textbf{-18.7} \\
 & & (C) + MoCrop (Attention, 224px) & \textbf{87.5} & 15.5 & \textbf{+0.2} & 0.0 \\
\bottomrule
\end{tabular}%
}
\end{table*}

\subsection{Experimental Setup}
\label{subsec:setup}
We evaluate MoCrop on the official UCF101 split. Following standard practice \cite{wu2018compressed}, we sample three frames per video for training and use typical data augmentation. At inference, we freeze the model weights and evaluate two main settings:
\textbf{Baseline:} The model trained on UCF101 is evaluated on the standard test set with a 224px input.
\textbf{MoCrop:} The same model is evaluated after each video is pre-processed by MoCrop. We test two variants:
\begin{itemize}
    \item \textbf{Efficiency Setting:} MoCrop crops the video, which is then resized to a smaller resolution (192px) to reduce FLOPs while aiming to maintain or improve accuracy.
    \item \textbf{Attention Setting:} MoCrop crops the video, which is resized back to the original resolution (224px). This isolates the accuracy improvement gained by focusing on salient regions, without changing the computational cost.
\end{itemize}

\subsection{Performance and Efficiency Analysis}

Table~\ref{tab:main_results_corrected} demonstrates the powerful dual advantages of MoCrop as a plug-and-play module for various modern backbones. In the Attention Setting, where FLOPs are held constant, MoCrop consistently boosts performance by forcing the model to concentrate on motion-rich areas. The most significant gain is seen with {ResNet-50~\cite{he2016deep}, which improves by a remarkable +3.5\% Top-1 accuracy. Other models like EfficientNet-B1~\cite{tan2019efficientnet} (+2.4\%) and MobileNet-V3~\cite{howard2019searching} (+1.9\%) also show substantial improvements.

In the Efficiency Setting, MoCrop proves its ability to reduce computational costs and often enhances accuracy. For instance, ResNet-50 sees a +2.4\% accuracy increase while FLOPs are cut by a massive 26.5\%. Even for the heavy Swin-B Transformer~\cite{liu2022video}, where accuracy gains are harder to achieve, our method reduces the computational load by 18.7\% with only a negligible 0.1\% accuracy dip. These results confirm that MoCrop offers an exceptional and highly flexible trade-off between accuracy and efficiency.

\subsection{Ablation Studies}
\label{sec:ablation}
To validate the contributions of the internal components of MoCrop and its effectiveness as a general module, we present a series of ablation studies in Tables~\ref{tab:component_ablation} and \ref{tab:comparisons}. The studies are organized into three parts: (a) an incremental analysis of our module's components, (b) a comparison against standard cropping methods, and (c) an application of MoCrop to enhance a prior state-of-the-art method.

\runinhead{ Incremental Component Analysis}
Using ResNet-18~\cite{he2016deep} as a base, we find that each component of MoCrop contributes to the final performance (Table~\ref{tab:component_ablation}). MGS is essential as the pipeline cannot run without it. Further incorporating MD or MCS individually yields substantial improvements, and the full pipeline combining all three components (MGS + MD + MCS) achieves the best results, confirming that all parts of our design are integral to its success. Importantly, the preprocessing cost column reveals the remarkable computational efficiency: even MGS alone incurs only 0.121 MOps, and adding MD or MCS reduces this to just 0.022 and 0.031 MOps respectively. The full pipeline achieves the lowest cost at 0.021 MOps—over 70,000× smaller than the model inference cost of 1.54 GFLOPs (1540 MOps)—demonstrating truly negligible overhead.

\begin{table}[t]
\centering
\caption{Component ablation study on ResNet-18. The Overhead column shows the negligible computational cost of MGS preprocessing (place\_blocks + search\_box) in Million Operations (MOps), compared to model inference at 1.54 GFLOPs (1540 MOps).}
\label{tab:component_ablation}
\resizebox{\linewidth}{!}{%
\begin{tabular}{@{}lccccc@{}}
\toprule
\textbf{Setup} & \textbf{MGS} & \textbf{MD} & \textbf{MCS} & \textbf{GFLOPs} & \textbf{Overhead (MOps)} \\
\midrule
Baseline (No MoCrop) & -- & -- & -- & 1.82 & 0 \\
w/o MGS (Pipeline cannot run) & -- & -- & -- & -- & -- \\
+ MGS & \checkmark & -- & -- & 1.54 & 0.121 \\
+ MGS + MD & \checkmark & \checkmark & -- & 1.54 & 0.022 \\
+ MGS + MCS & \checkmark & -- & \checkmark & 1.54 & 0.031 \\
\textbf{+ MGS + MD + MCS (Full)} & \textbf{\checkmark} & \textbf{\checkmark} & \textbf{\checkmark} & \textbf{1.54} & \textbf{0.021} \\
\bottomrule
\end{tabular}
}
\end{table}

\begin{table}[t]
\centering
\caption{Performance comparisons on ResNet-18 and CoViAR. (a) Comparison with standard cropping strategies. (b) Enhancing CoViAR with MoCrop on UCF101 (I-frames only).}
\label{tab:comparisons}
\resizebox{0.85\linewidth}{!}{%
\begin{tabular}{@{}lcc@{}}
\toprule
\textbf{Method} & \textbf{Acc. (\%)} & \textbf{GFLOPs} \\
\midrule
\multicolumn{3}{l}{\textit{(a) Cropping Strategies on ResNet-18}} \\
Baseline & 79.3 & 1.82 \\
Random Crop (90\%)   & 80.4 & 1.73 \\
Center Crop (90\%)   & 80.2 & 1.73 \\
\textbf{MoCrop (Adaptive)} & \textbf{81.4} & \textbf{1.54} \\
\midrule
\multicolumn{3}{l}{\textit{(b) Enhancing CoViAR (UCF101, I-frames only)}} \\
CoViAR Baseline (Reported in \cite{wu2018compressed}) & 87.7 & 11.6 \\
\textbf{+ MoCrop (Efficiency, 192px)} & 88.5 & \textbf{8.5} \\
\textbf{+ MoCrop (Attention, 224px)} & \textbf{89.2} & 11.6 \\
\bottomrule
\end{tabular}
}
\end{table}

\runinhead{ Comparison with Cropping Strategies}
MoCrop significantly outperforms standard cropping techniques across multiple crop ratios (Table~\ref{tab:comparisons}). At 90\% crop ratio, both Random and Center crops offer slight improvements over the baseline (80.4\% for both). However, as the crop ratio decreases to 50\%, these fixed strategies show severe degradation (73.5\% and 71.4\%). In contrast, our adaptive method achieves 81.4\% accuracy with an adaptive crop ratio, demonstrating superior motion-guided region selection over static approaches across all settings.

\runinhead{ Enhancing Prior Works}
To show its versatility, we applied MoCrop to CoViAR, a well-known compressed-domain method (Table~\ref{tab:comparisons}). The results are compelling. In the efficiency setting, MoCrop improves CoViAR's accuracy to 88.5\% while reducing its FLOPs from 11.6 GFLOPs to just 8.5 GFLOPs. In the attention setting, it boosts the accuracy to an impressive 89.2\%. This shows that MoCrop is not just a standalone module but can also serve as a powerful enhancer for other methods in the field.

\subsection{Qualitative Analysis}
\label{subsec:qualitative}

Fig.~\ref{fig:pipeline_complete} visualizes the complete MoCrop pipeline across six representative UCF101 action classes. Each row demonstrates how raw MVs are progressively refined through MD (percentile filtering), MCS (importance sampling), and MGS (grid aggregation and region search) to produce focused crops. The final comparison (column f) shows that MoCrop consistently identifies tighter, action-centric regions compared to fixed center crops, removing irrelevant background while preserving the actor.

\newcommand{\triptych}[4][draw,line width=0.6pt]{%
  \begin{subfigure}{0.24\textwidth}
    \centering
    \begin{tikzpicture}
      \node[inner sep=2pt, #1] (wrap) {%
        \begin{minipage}{0.98\linewidth}\centering
          \includegraphics[width=0.49\linewidth]{#2}\hfill
          \includegraphics[width=0.49\linewidth]{#3}
        \end{minipage}
      };
    \end{tikzpicture}
    \vspace{-1.6em}
    \subcaption{#4}
  \end{subfigure}%
}

\begin{figure*}[t]
\centering
\scriptsize 
\setlength{\tabcolsep}{1pt} 

\begin{tabular}{cccccc}
\textbf{(a) Raw MVs} & \textbf{(b) MD} & \textbf{(c) MCS} & 
\textbf{(d) MGS(G)} & \textbf{(e) MGS(S)} & \textbf{(f) Comparison} \\[2pt]

\includegraphics[width=0.15\textwidth]{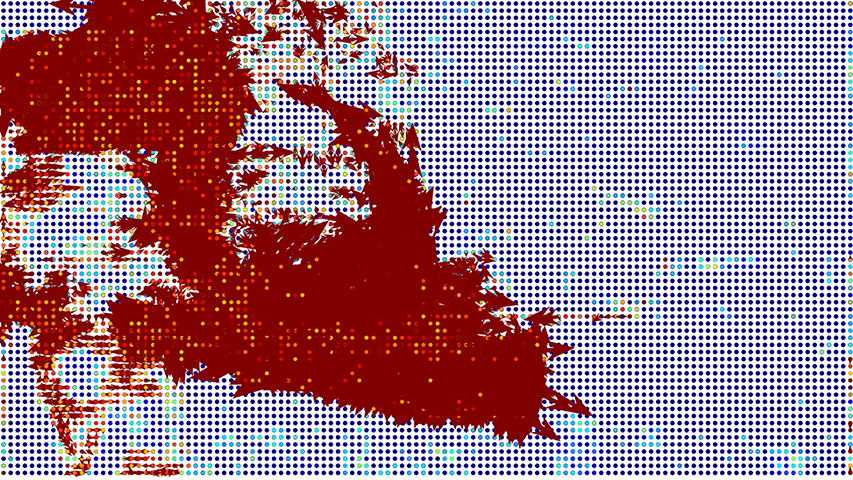} &
\includegraphics[width=0.15\textwidth]{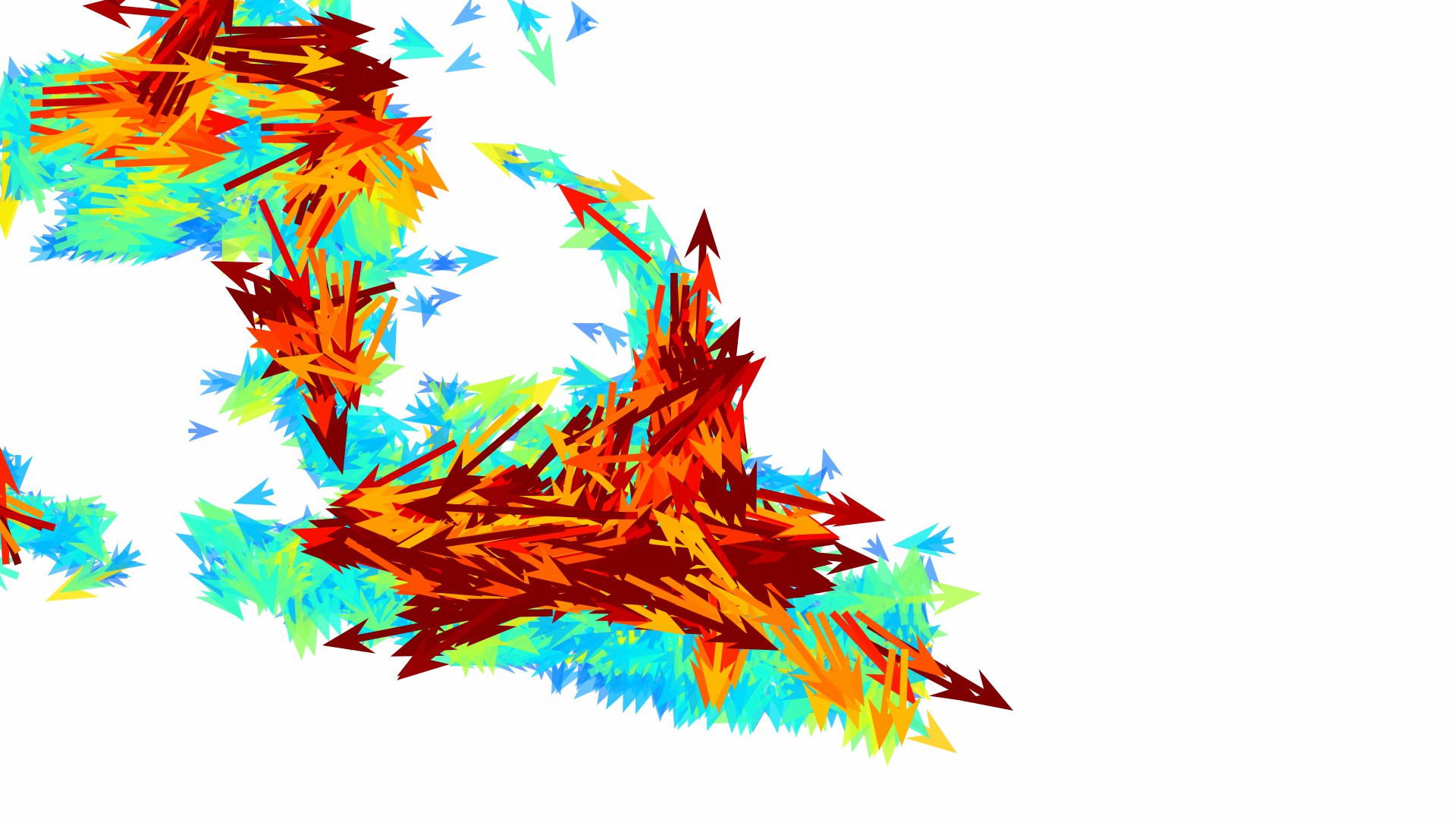} &
\includegraphics[width=0.15\textwidth]{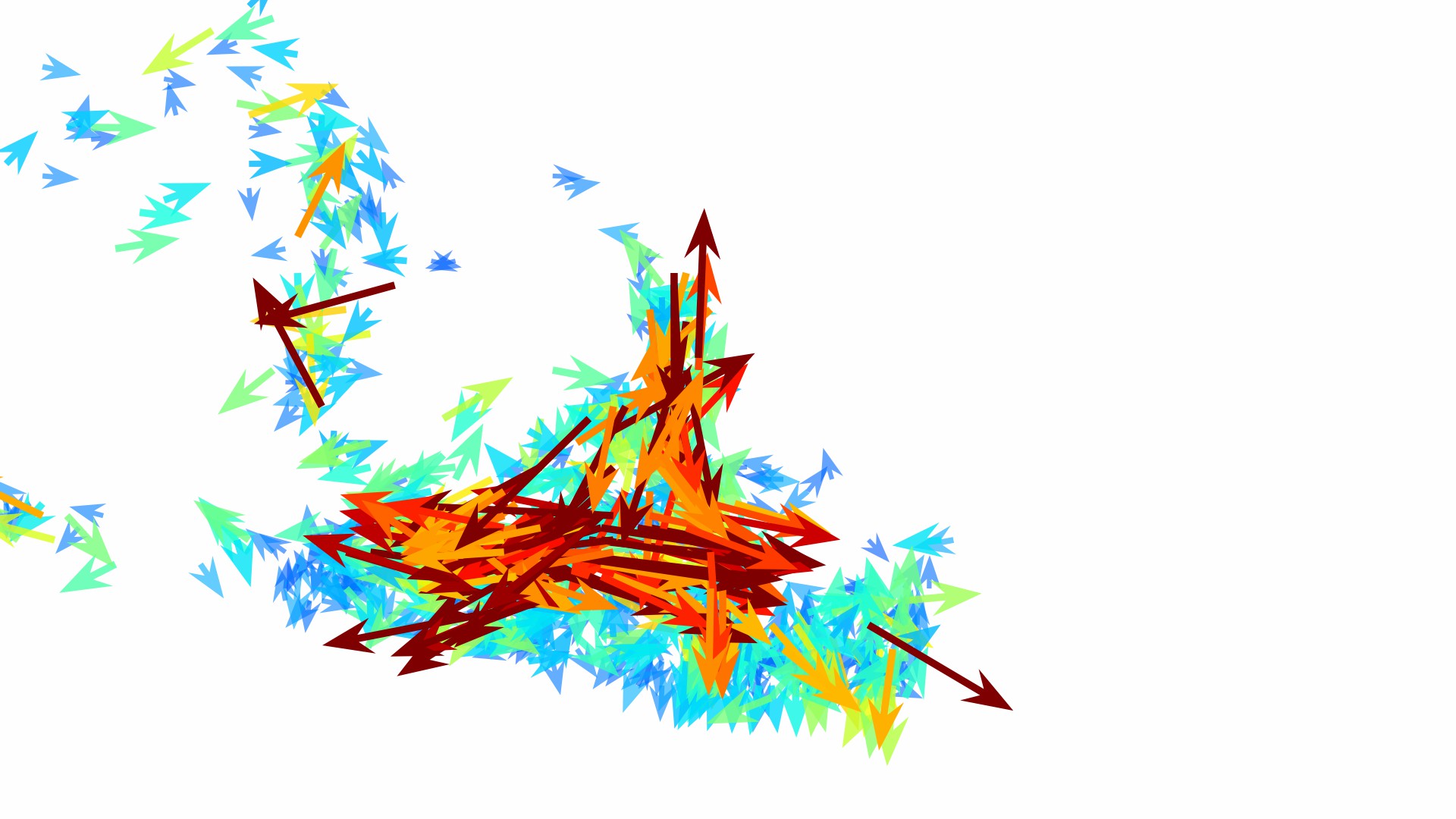} &
\includegraphics[width=0.15\textwidth]{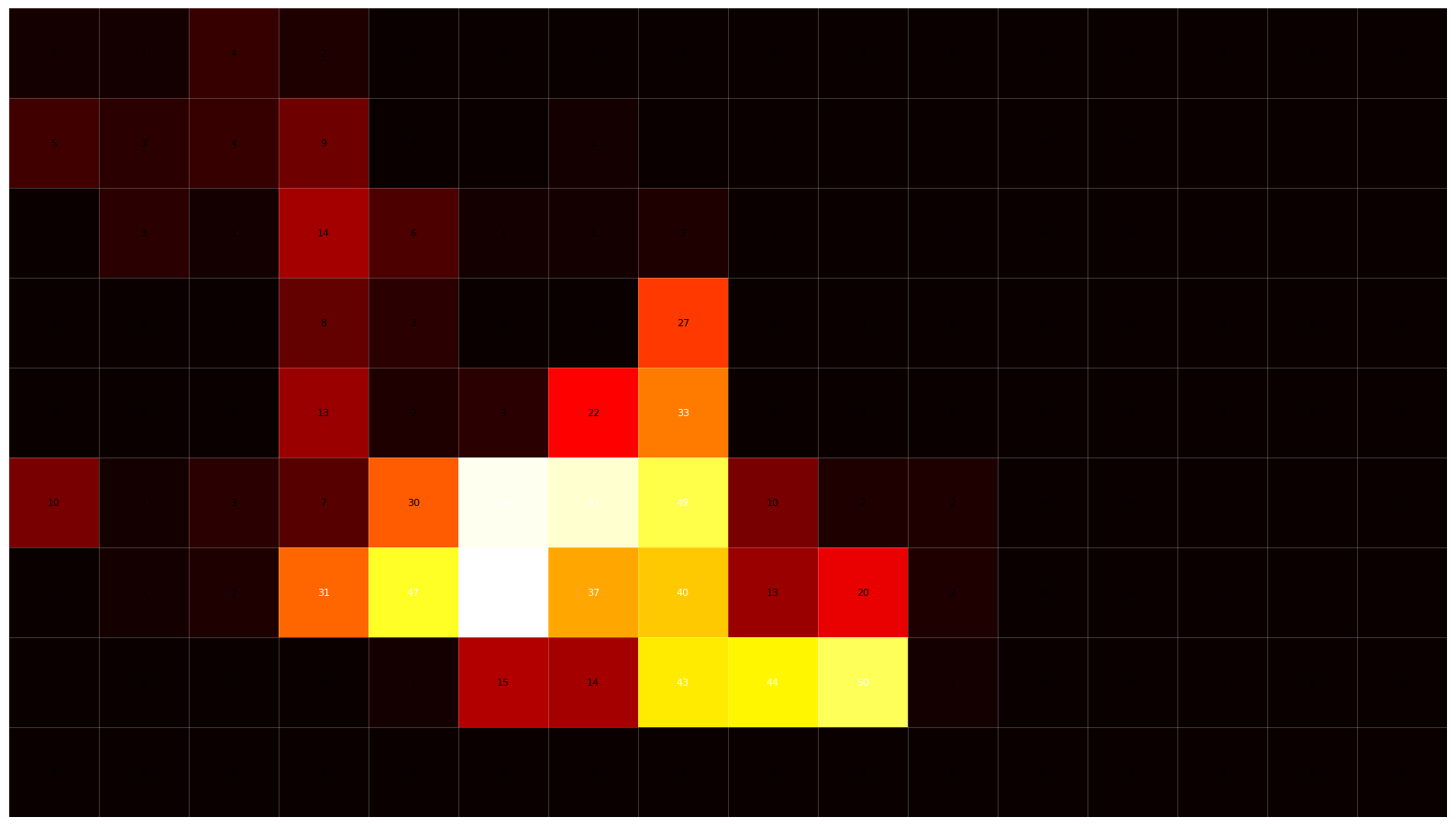} &
\includegraphics[width=0.15\textwidth]{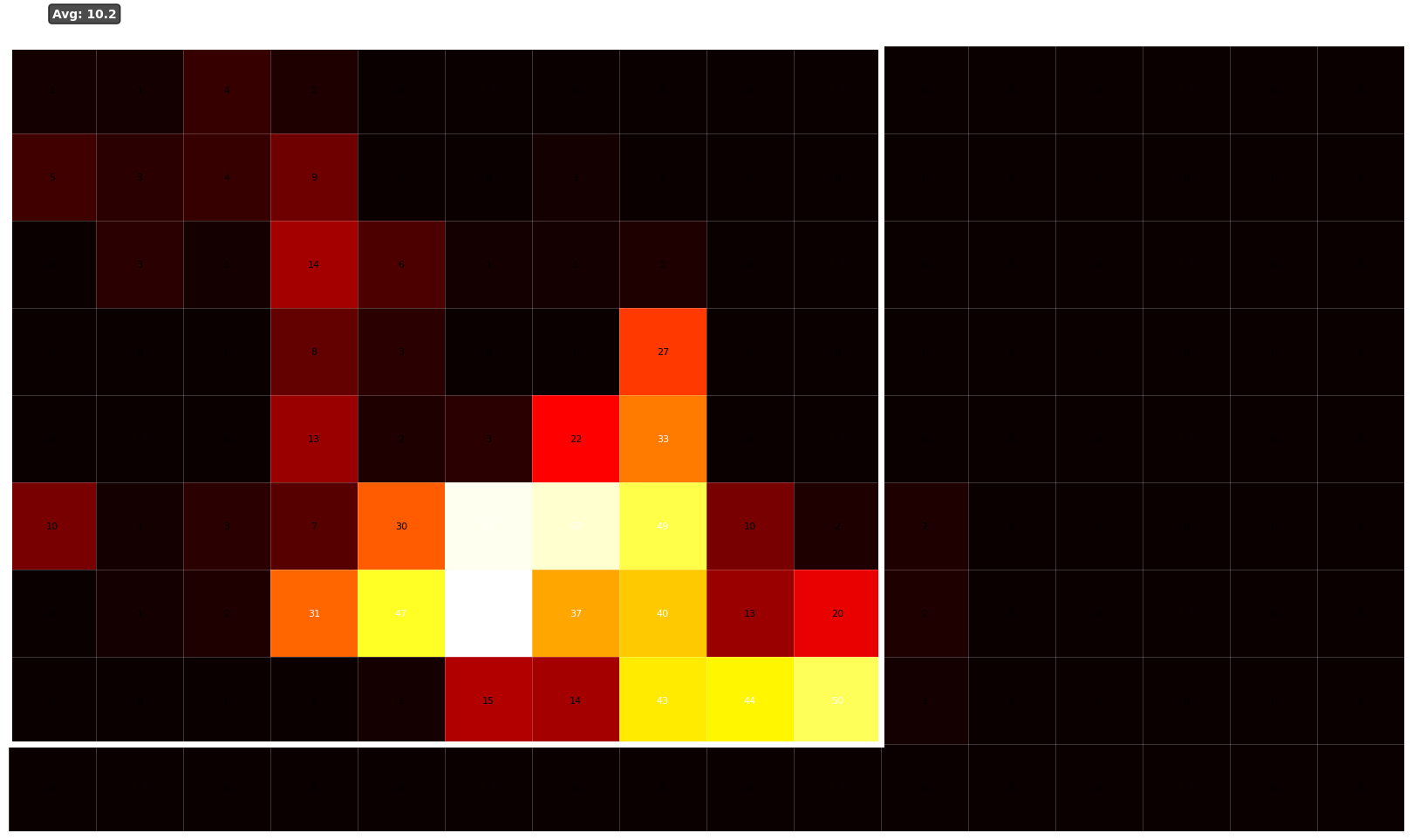} &
\includegraphics[width=0.15\textwidth]{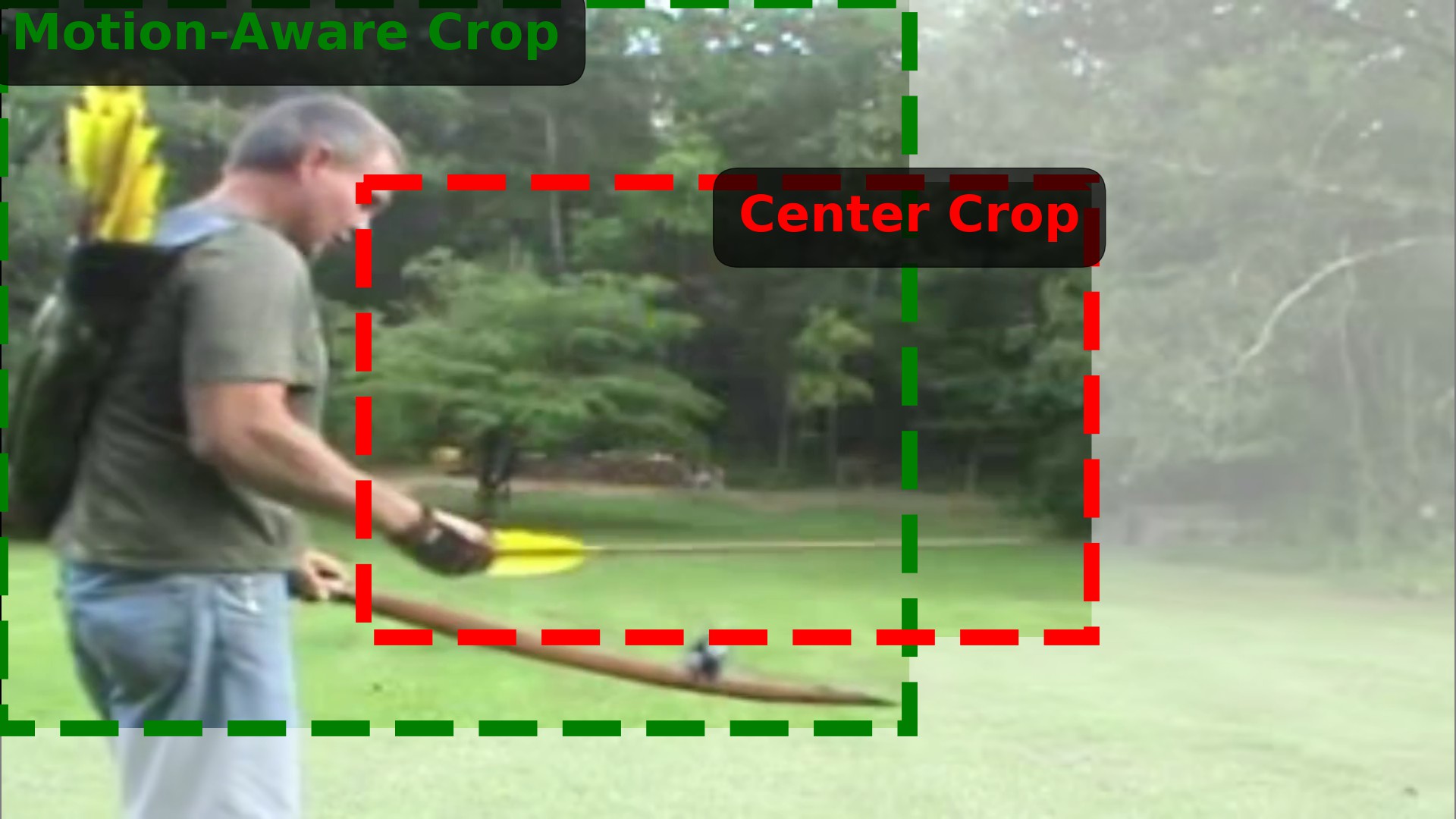} \\
\multicolumn{6}{l}{\textit{Archery}} \\[4pt]

\includegraphics[width=0.15\textwidth]{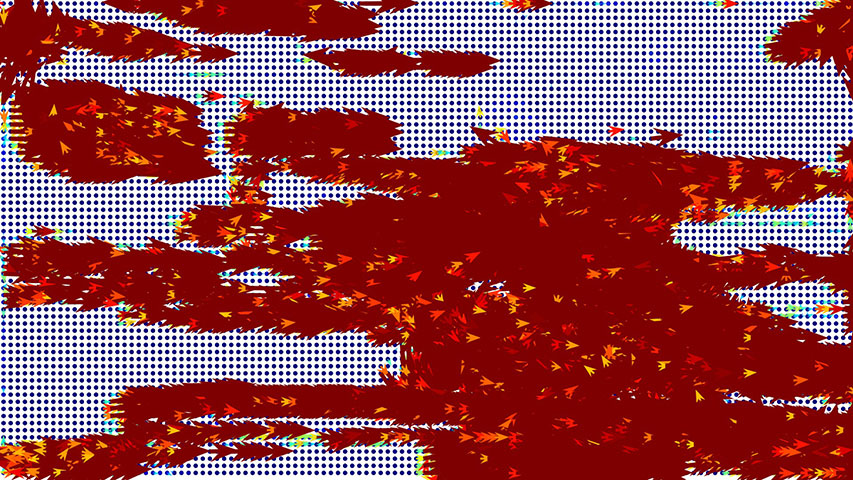} &
\includegraphics[width=0.15\textwidth]{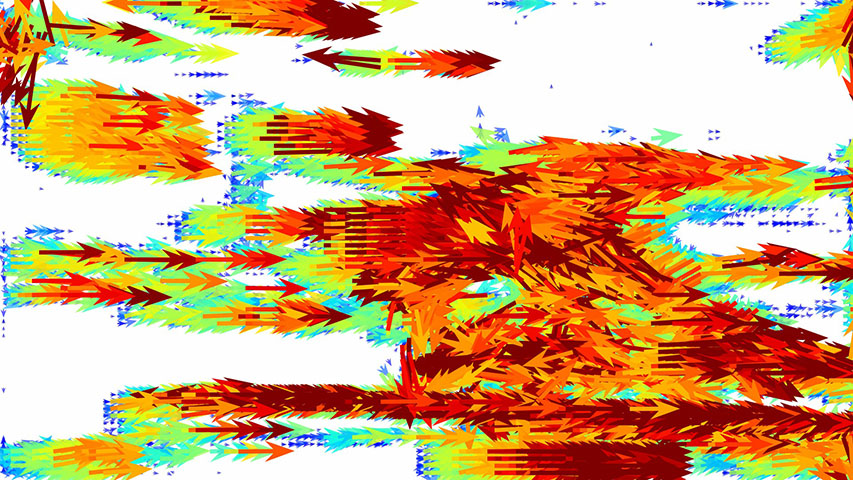} &
\includegraphics[width=0.15\textwidth]{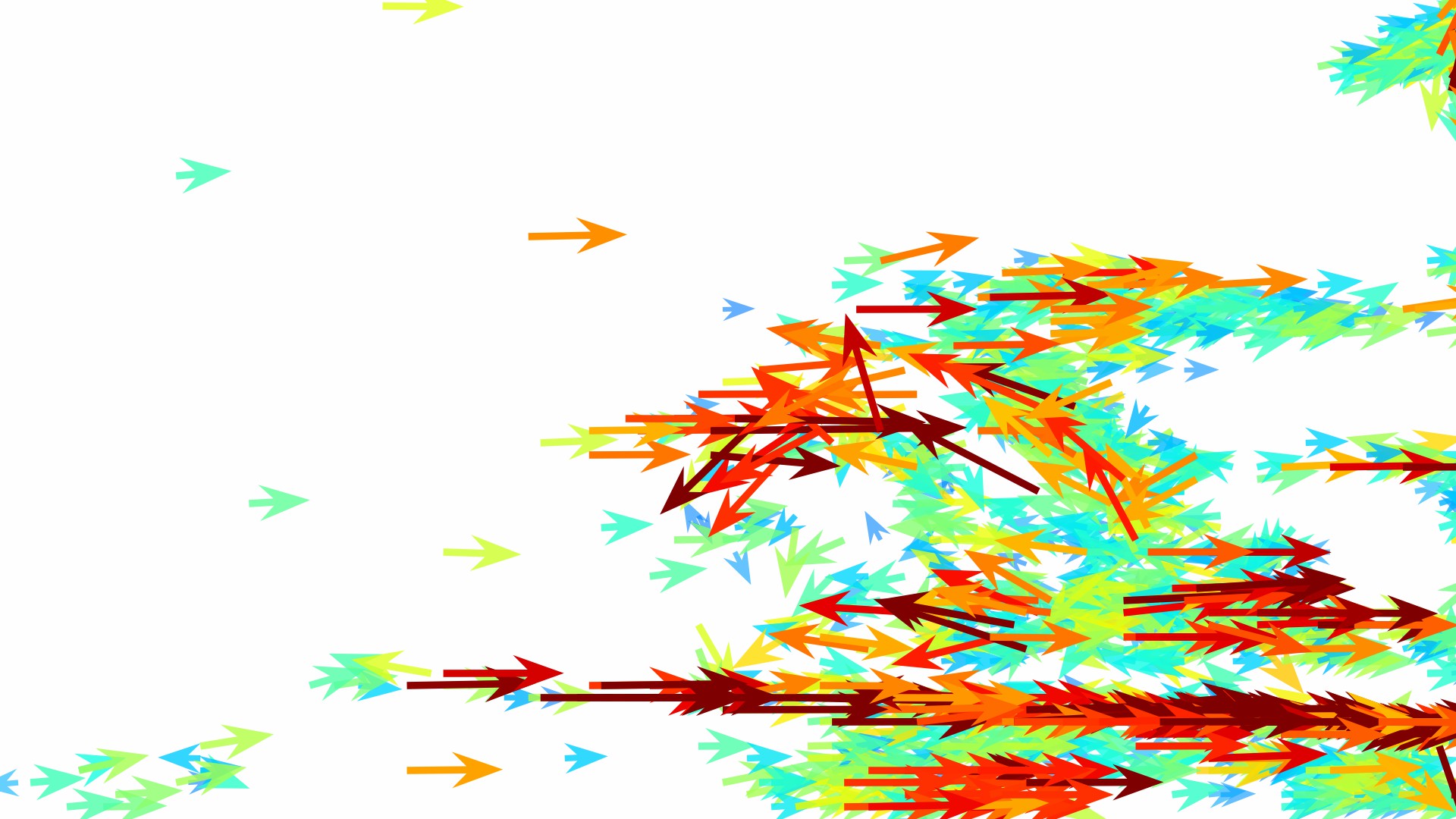} &
\includegraphics[width=0.15\textwidth]{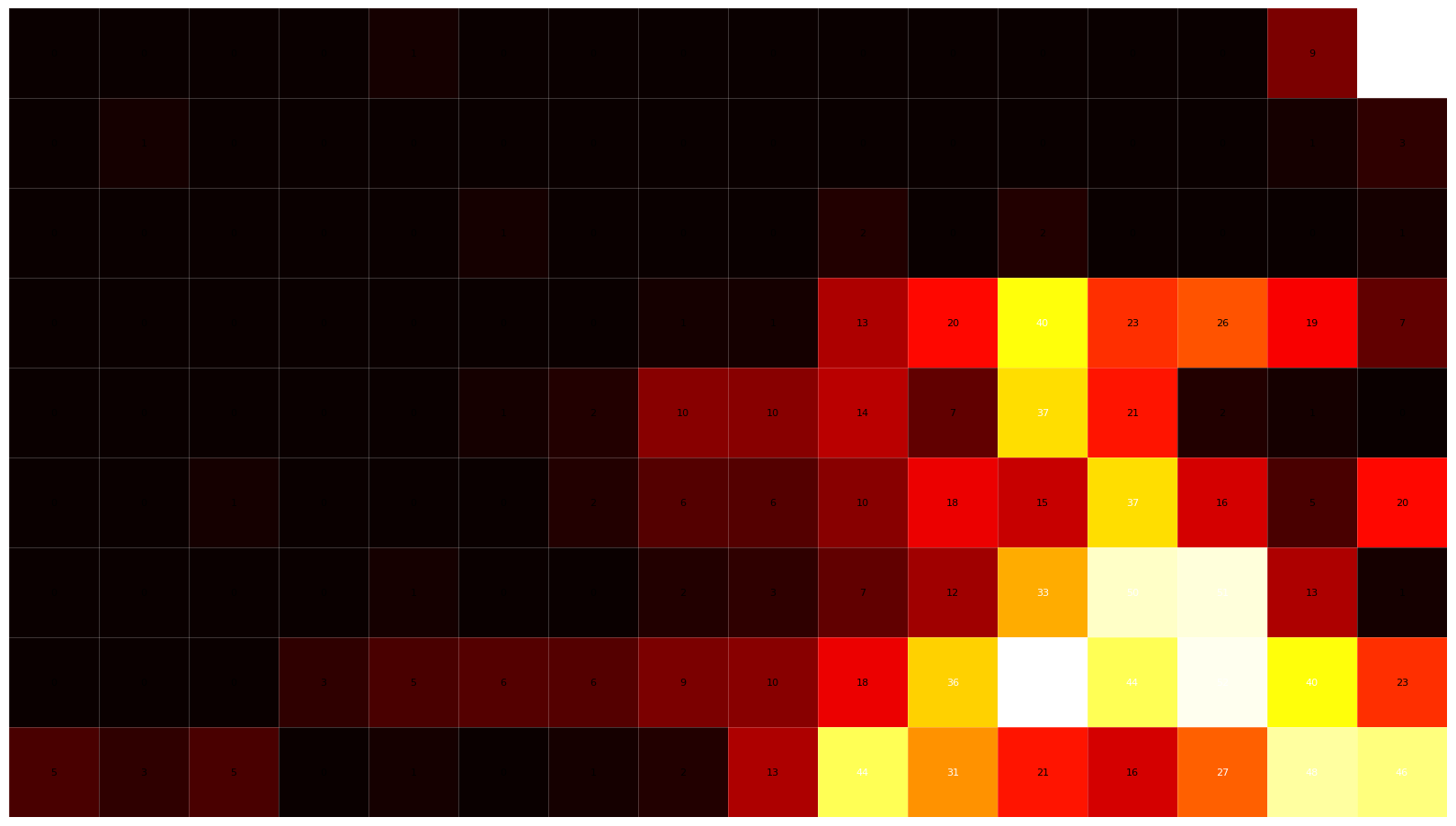} &
\includegraphics[width=0.15\textwidth]{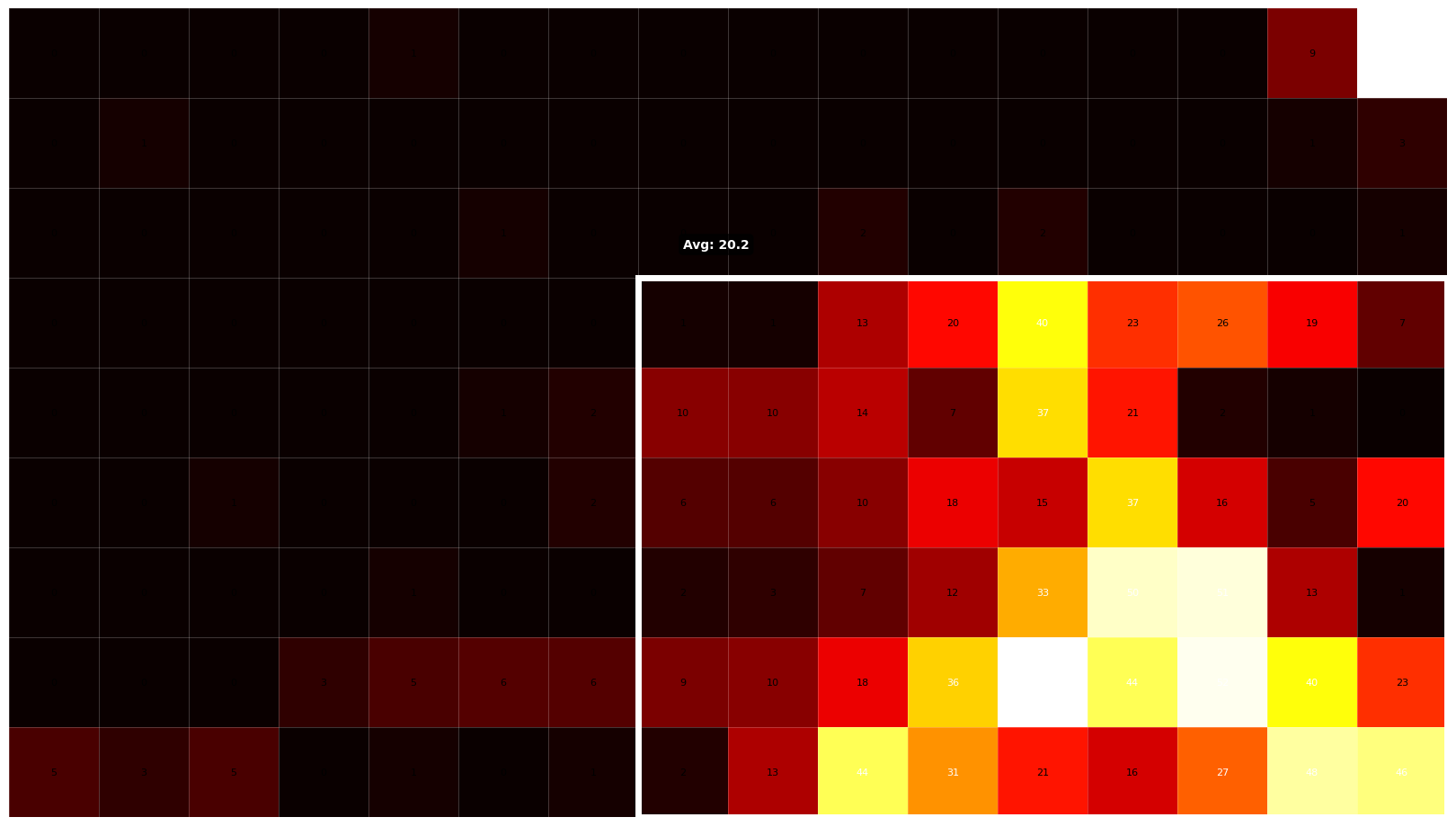} &
\includegraphics[width=0.15\textwidth]{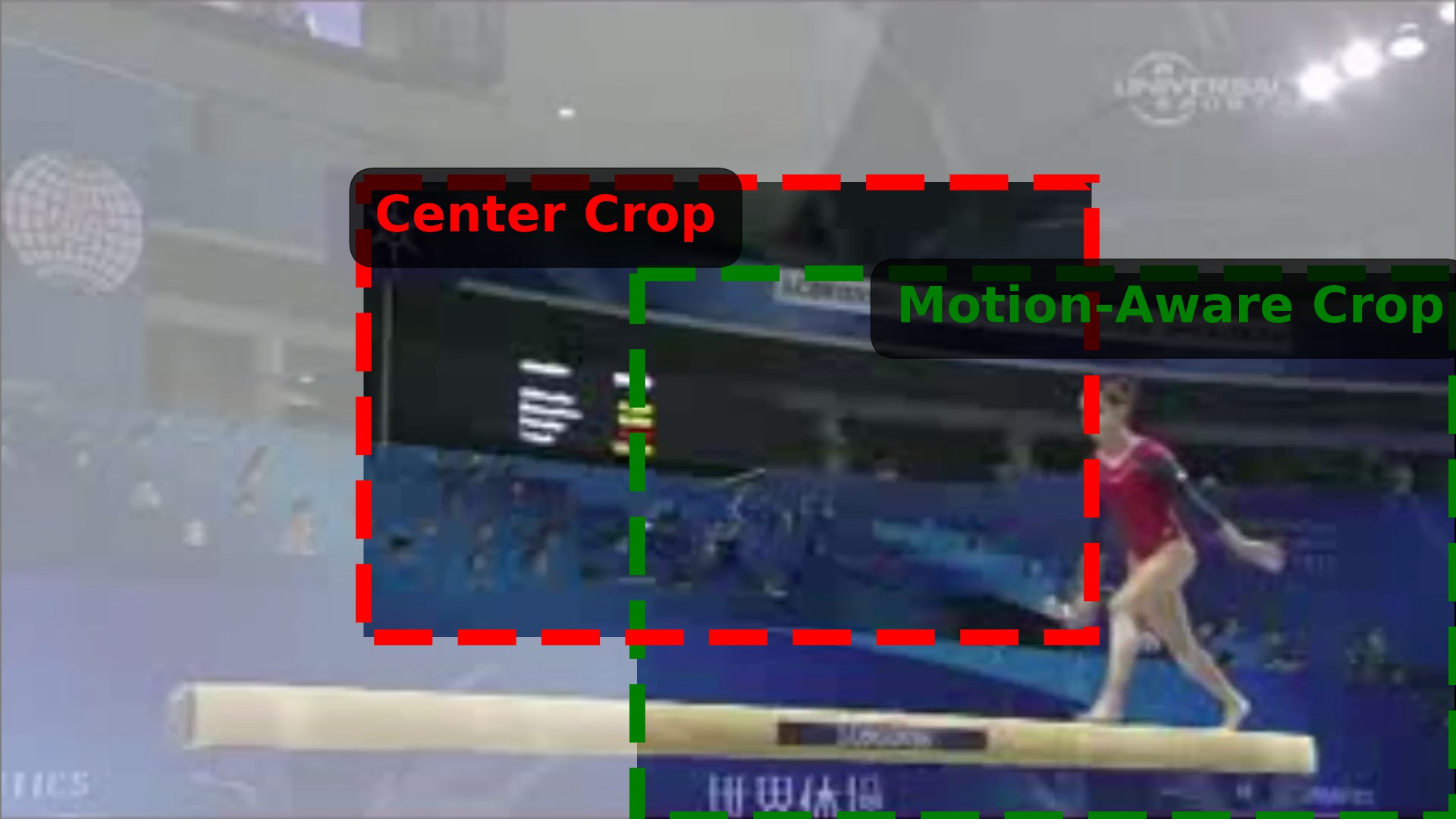} \\
\multicolumn{6}{l}{\textit{BalanceBeam}} \\[4pt]

\includegraphics[width=0.15\textwidth]{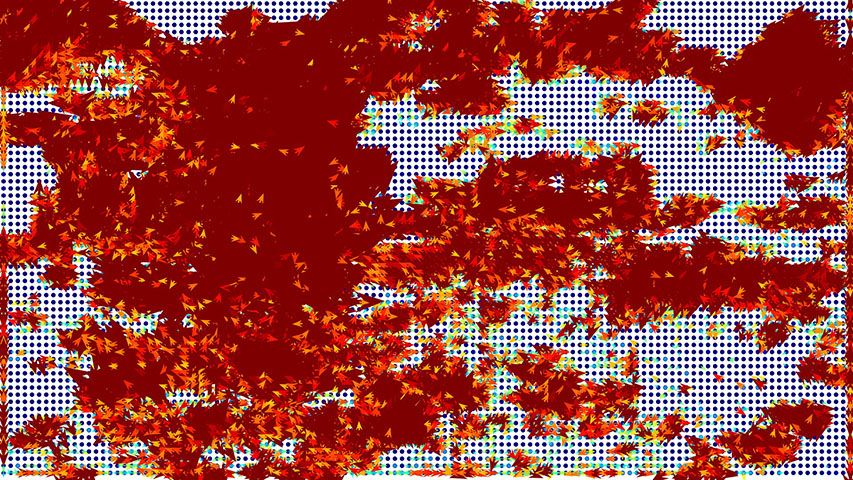} &
\includegraphics[width=0.15\textwidth]{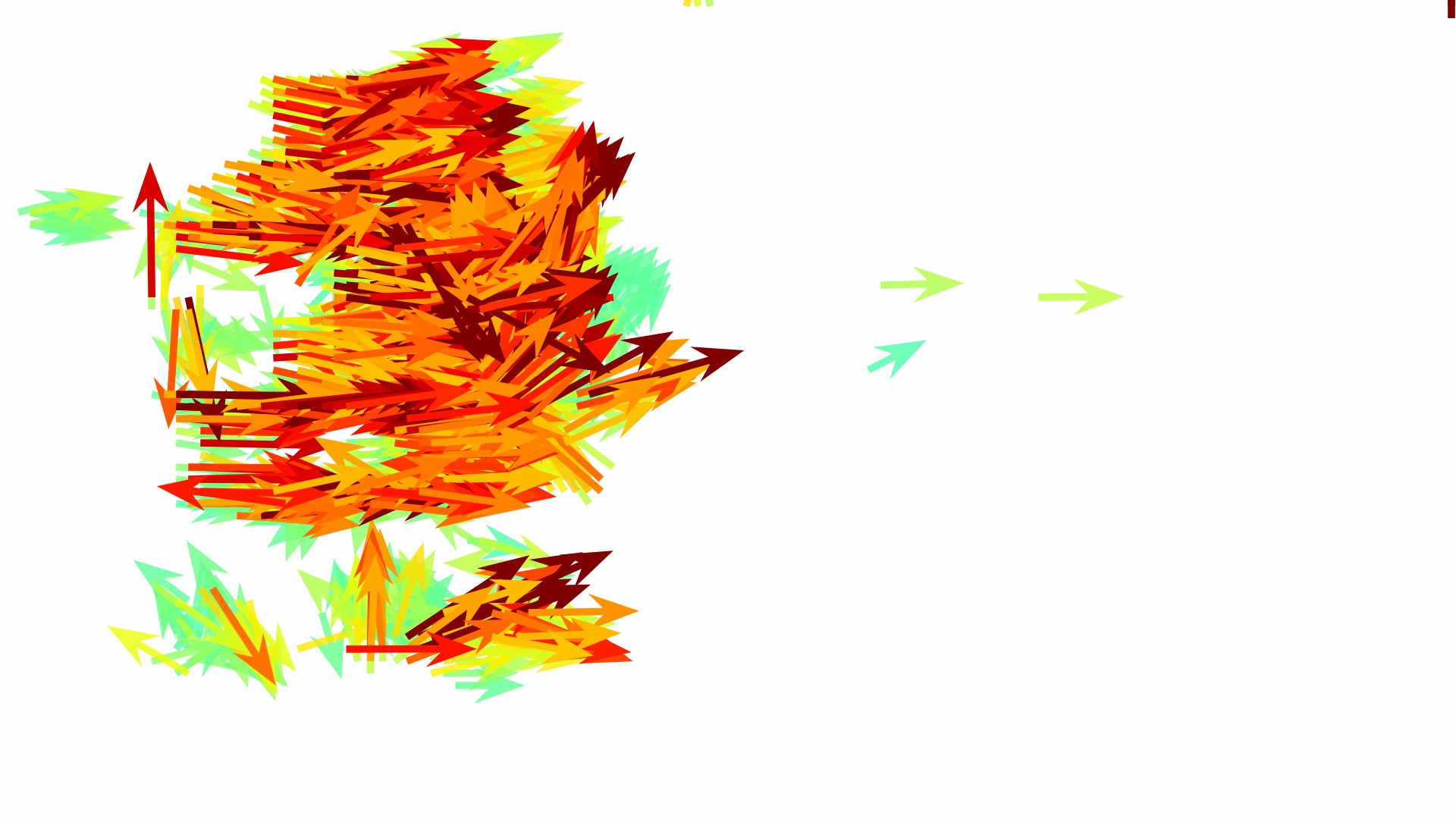} &
\includegraphics[width=0.15\textwidth]{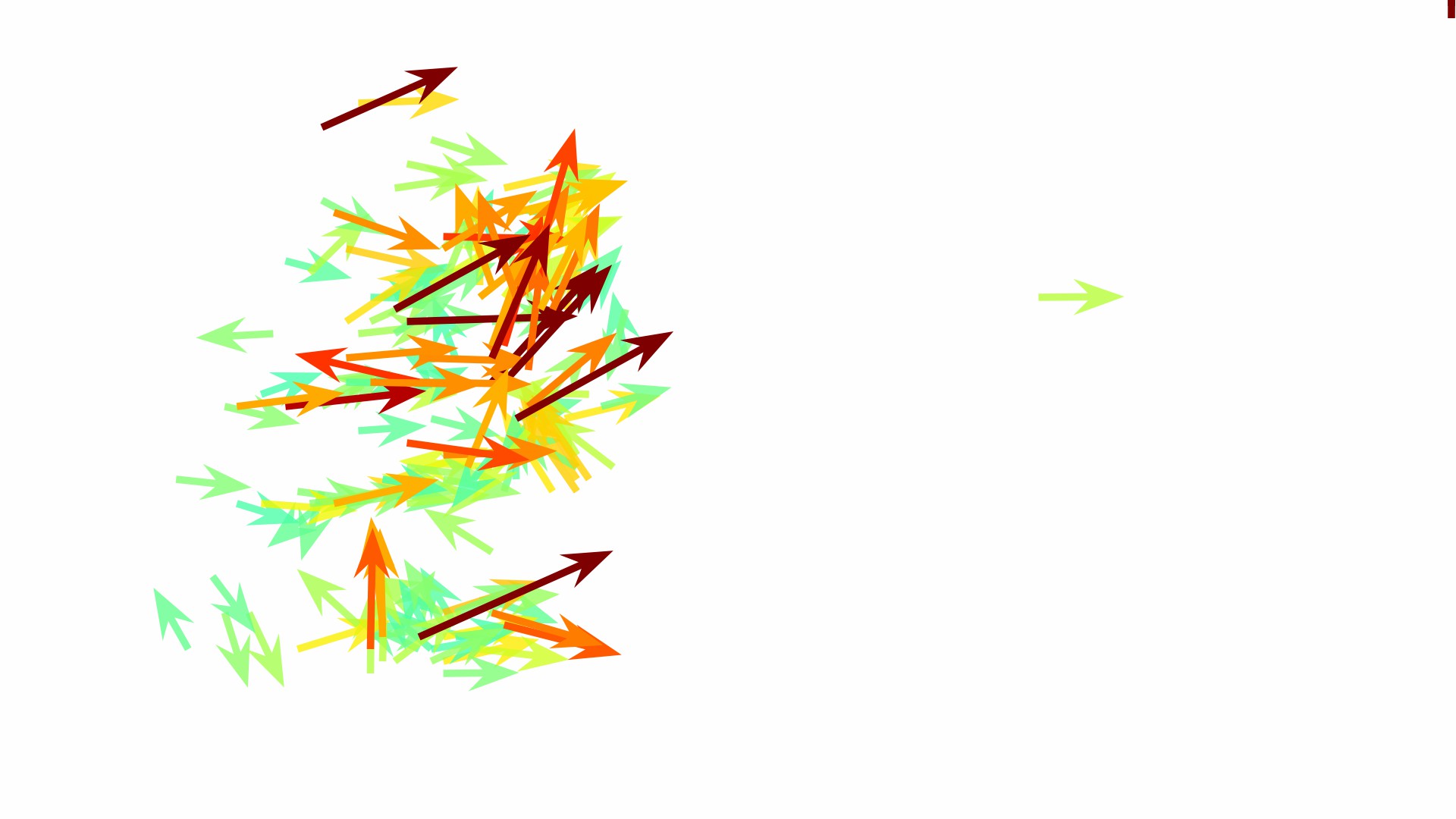} &
\includegraphics[width=0.15\textwidth]{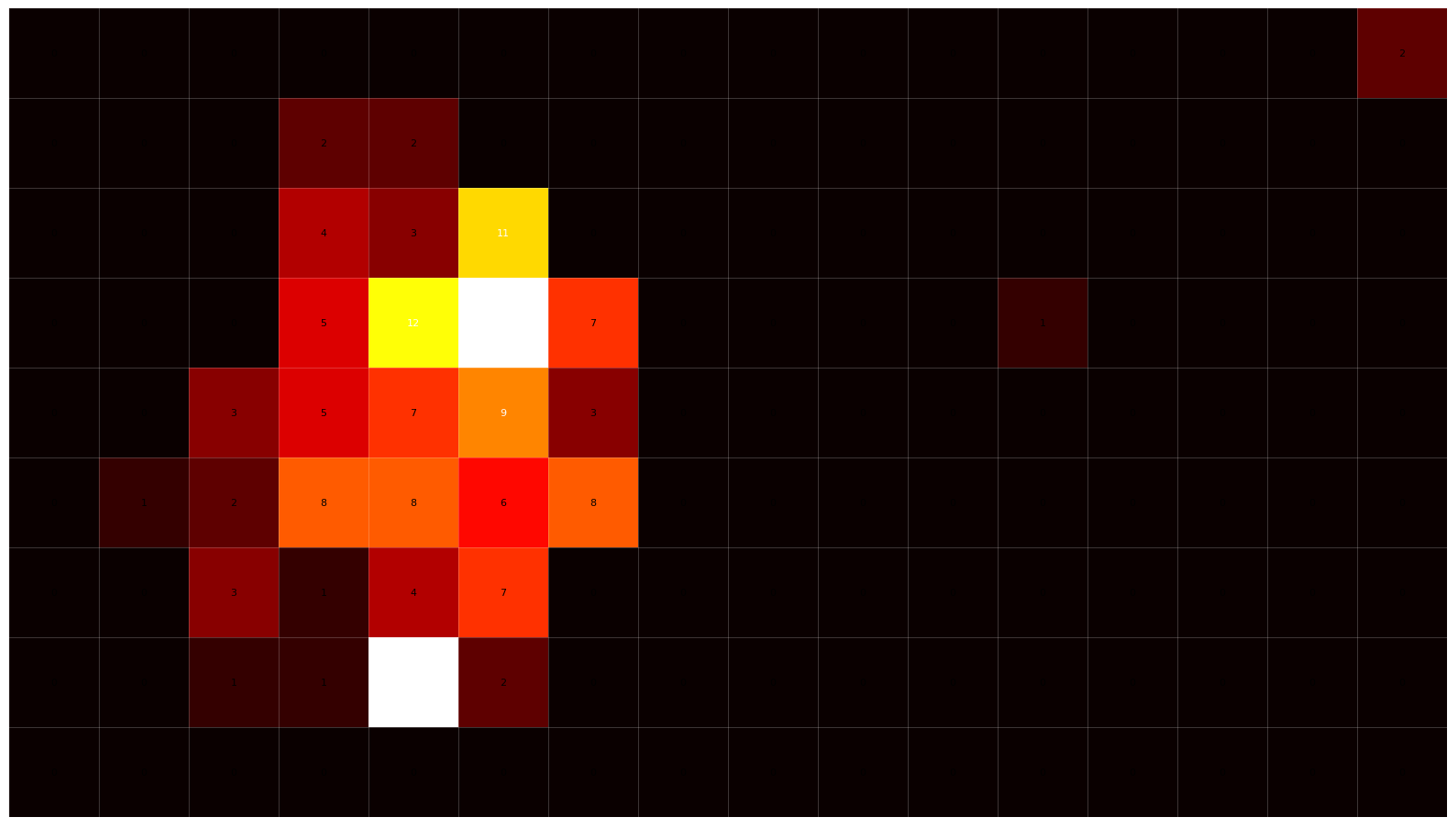} &
\includegraphics[width=0.15\textwidth]{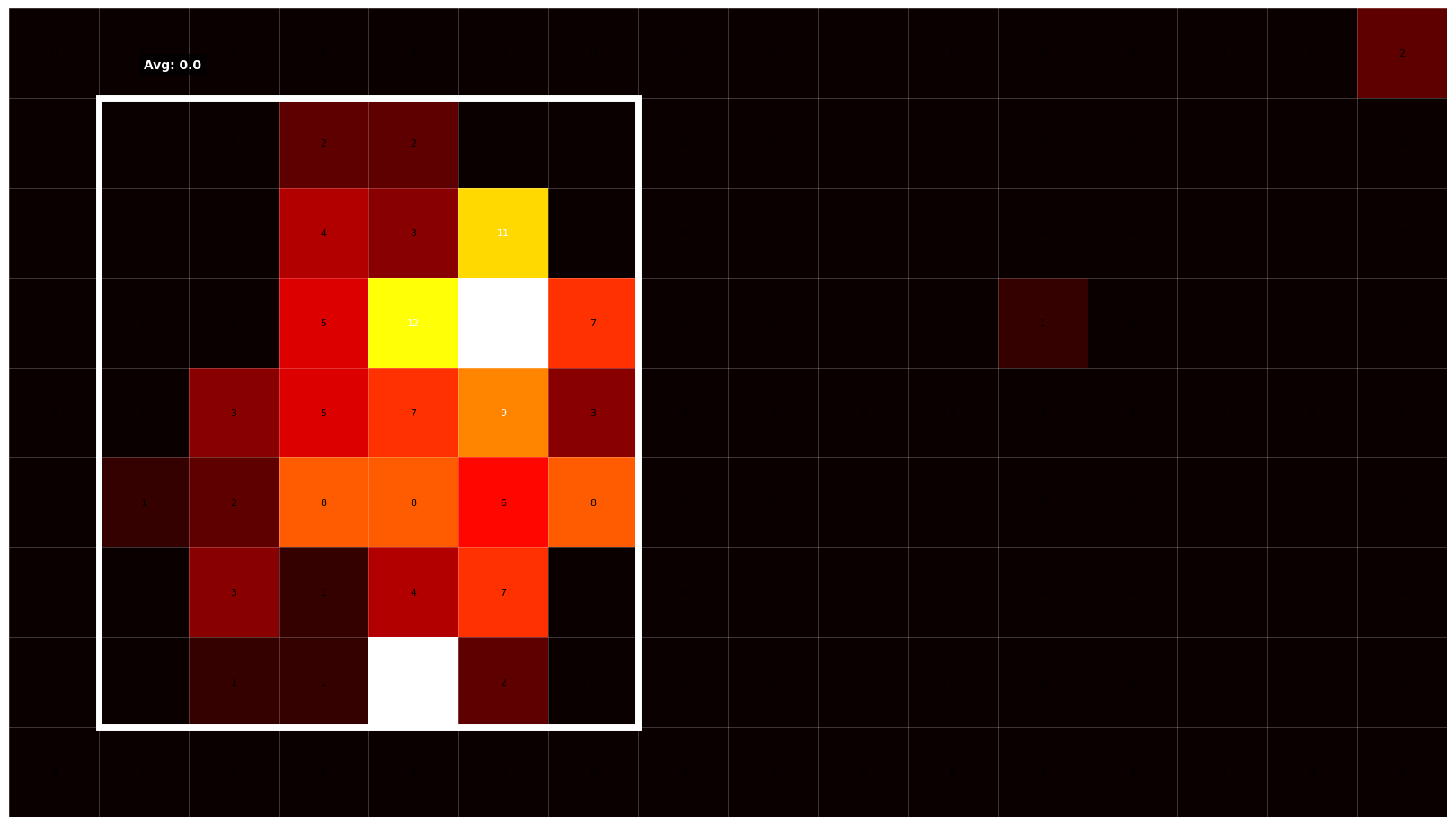} &
\includegraphics[width=0.15\textwidth]{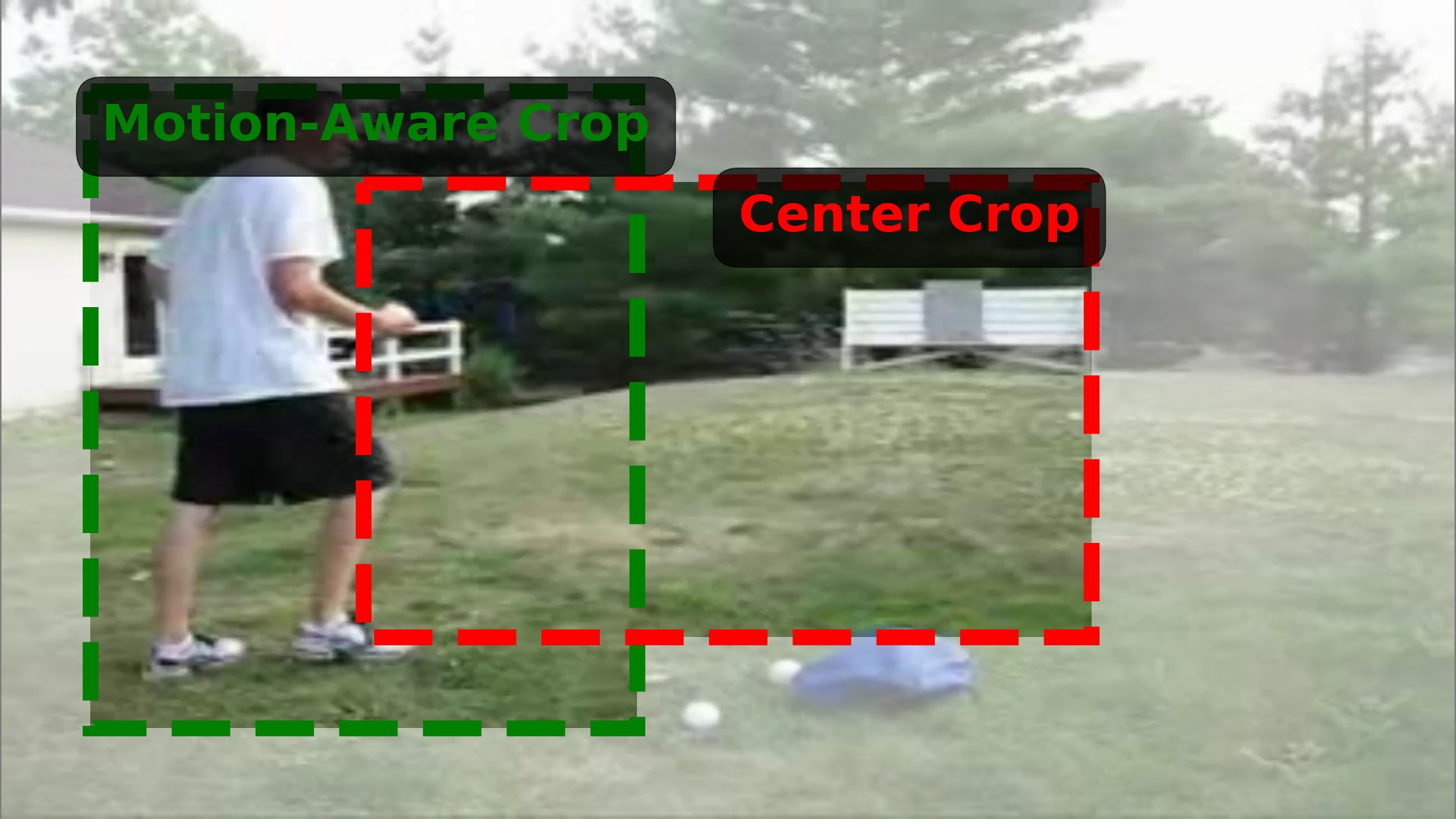} \\
\multicolumn{6}{l}{\textit{BaseballPitch}} \\[4pt]

\includegraphics[width=0.15\textwidth]{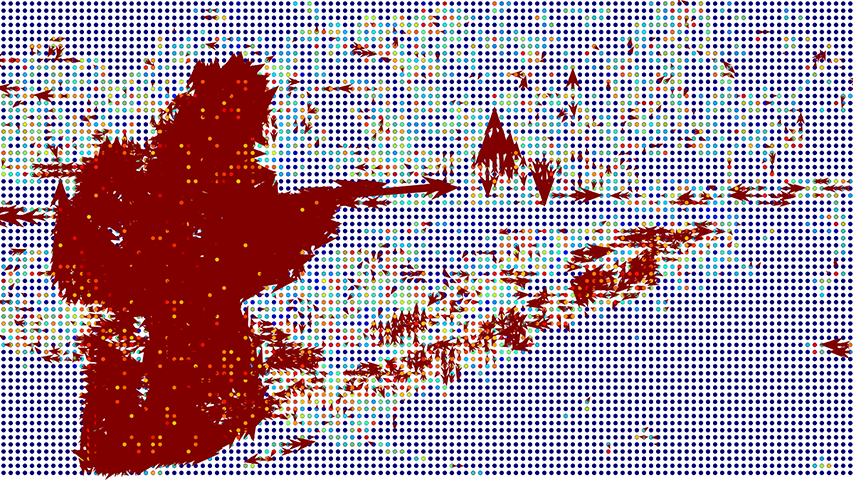} &
\includegraphics[width=0.15\textwidth]{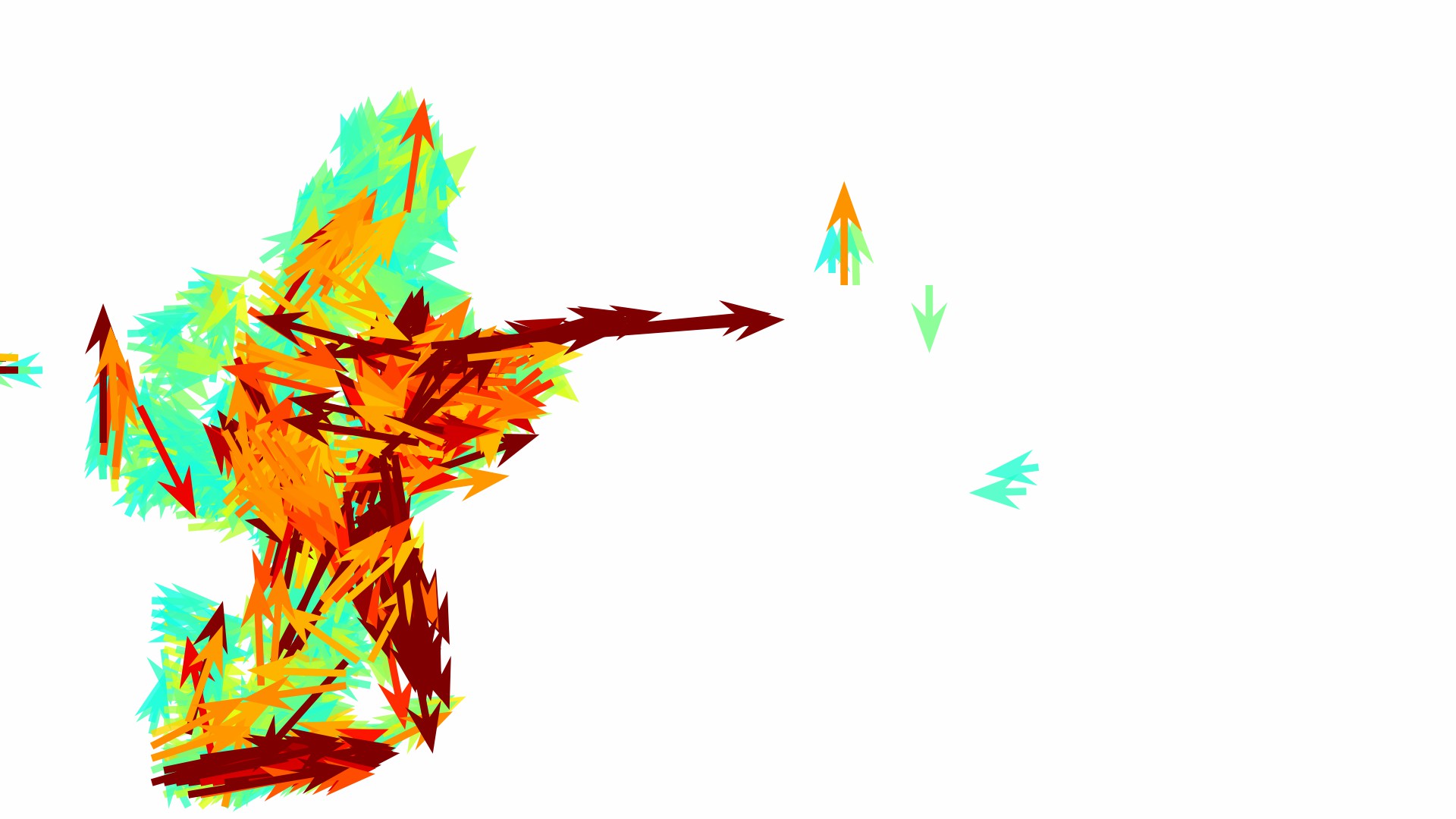} &
\includegraphics[width=0.15\textwidth]{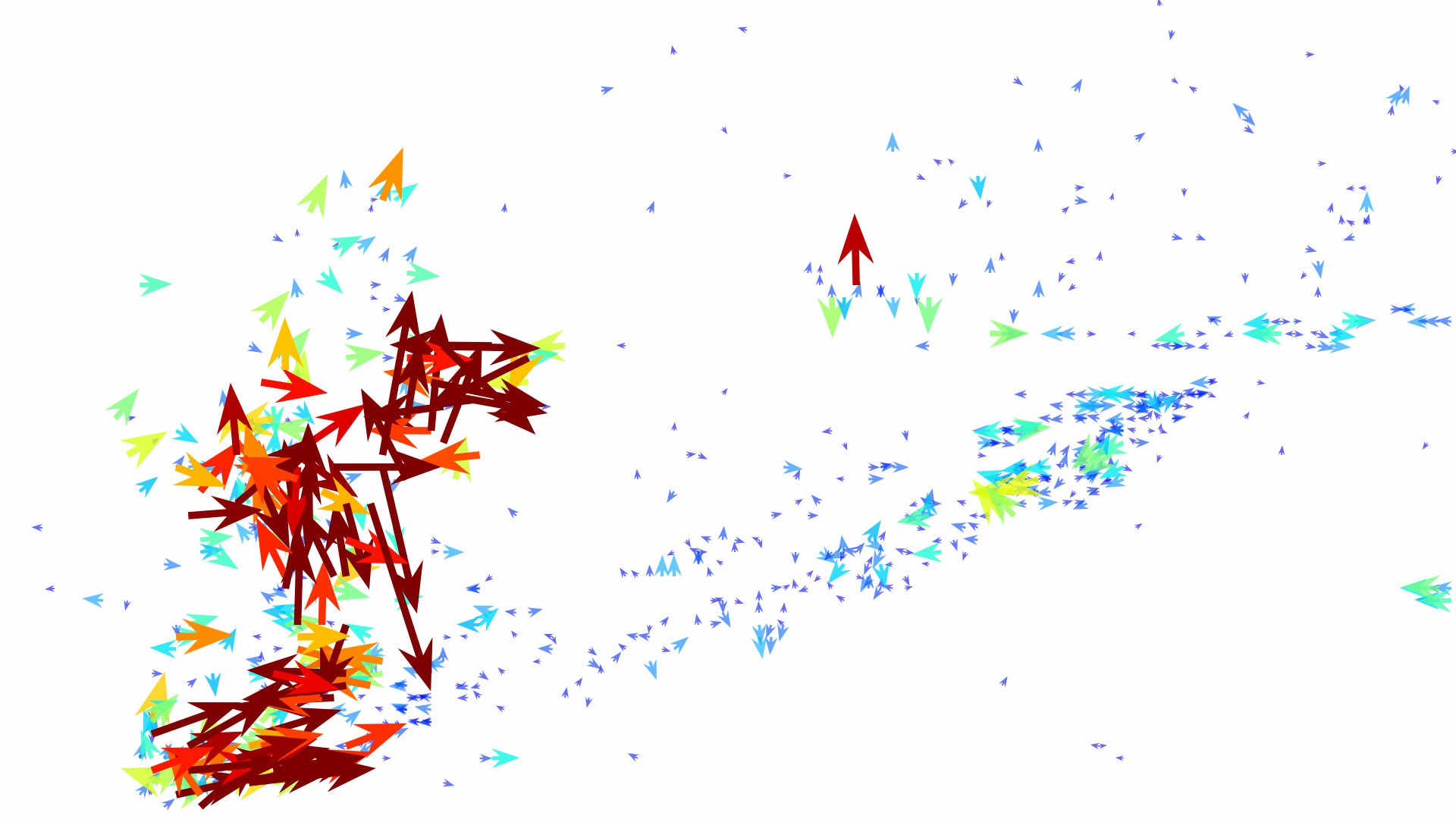} &
\includegraphics[width=0.15\textwidth]{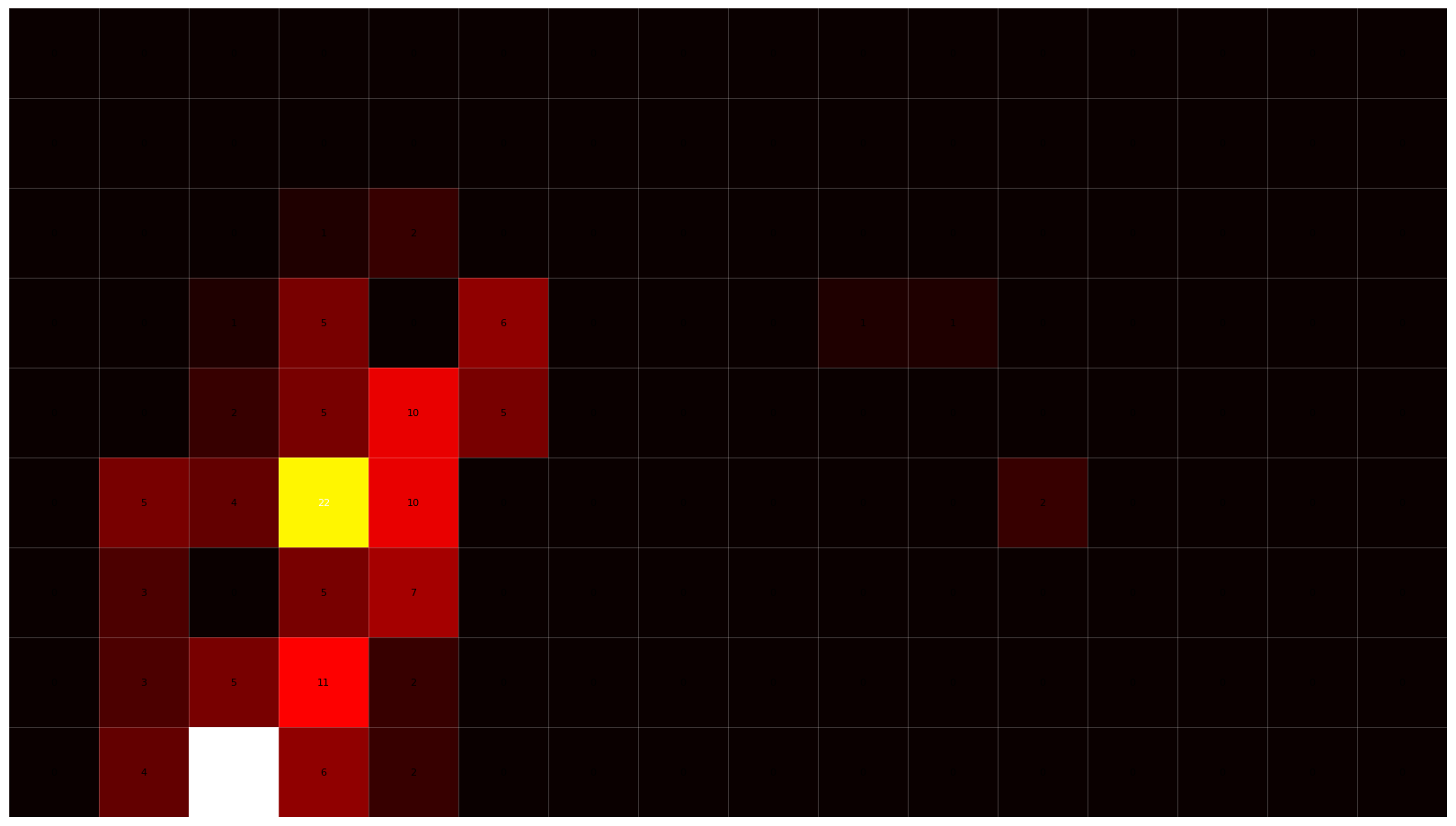} &
\includegraphics[width=0.15\textwidth]{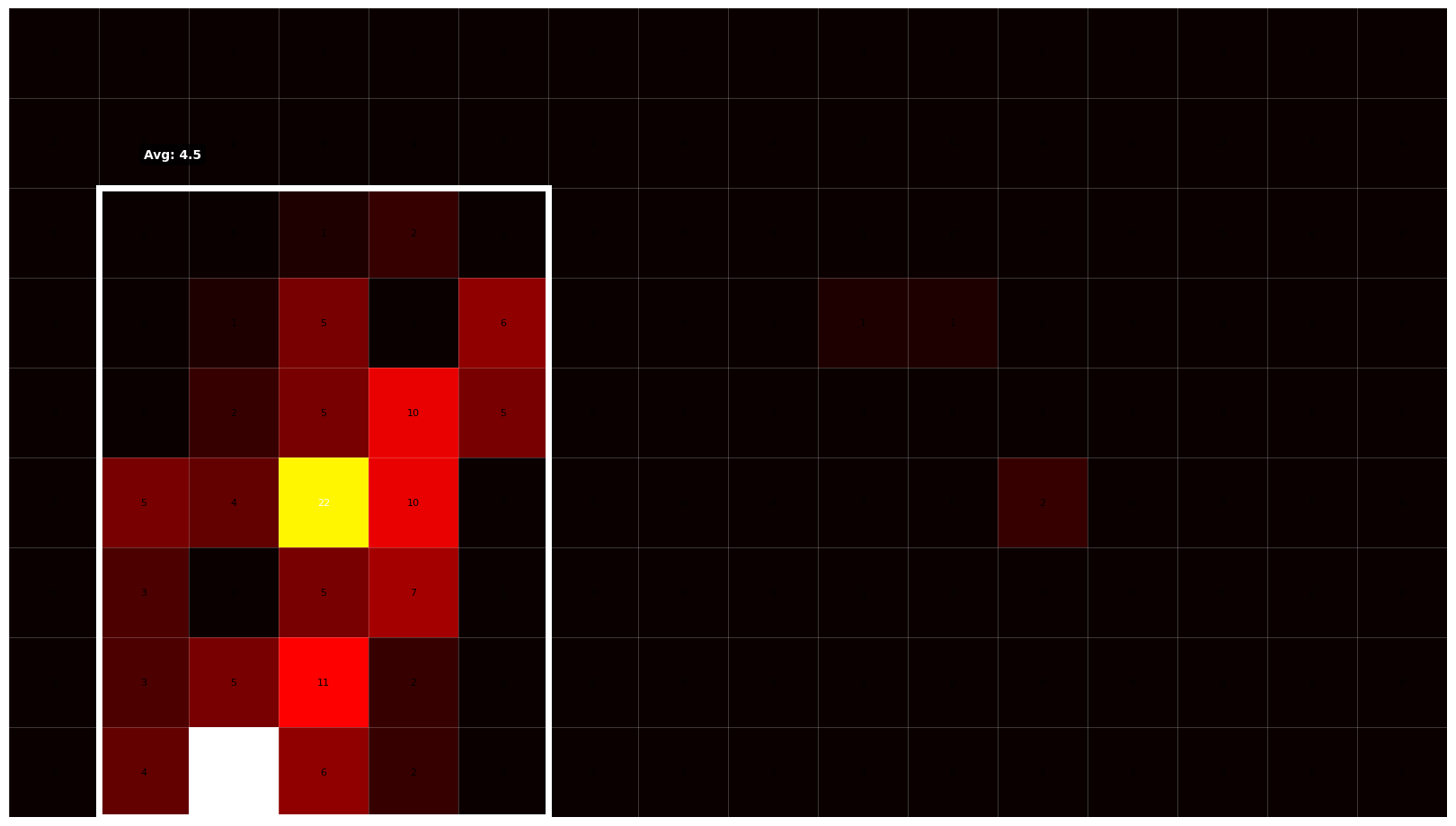} &
\includegraphics[width=0.15\textwidth]{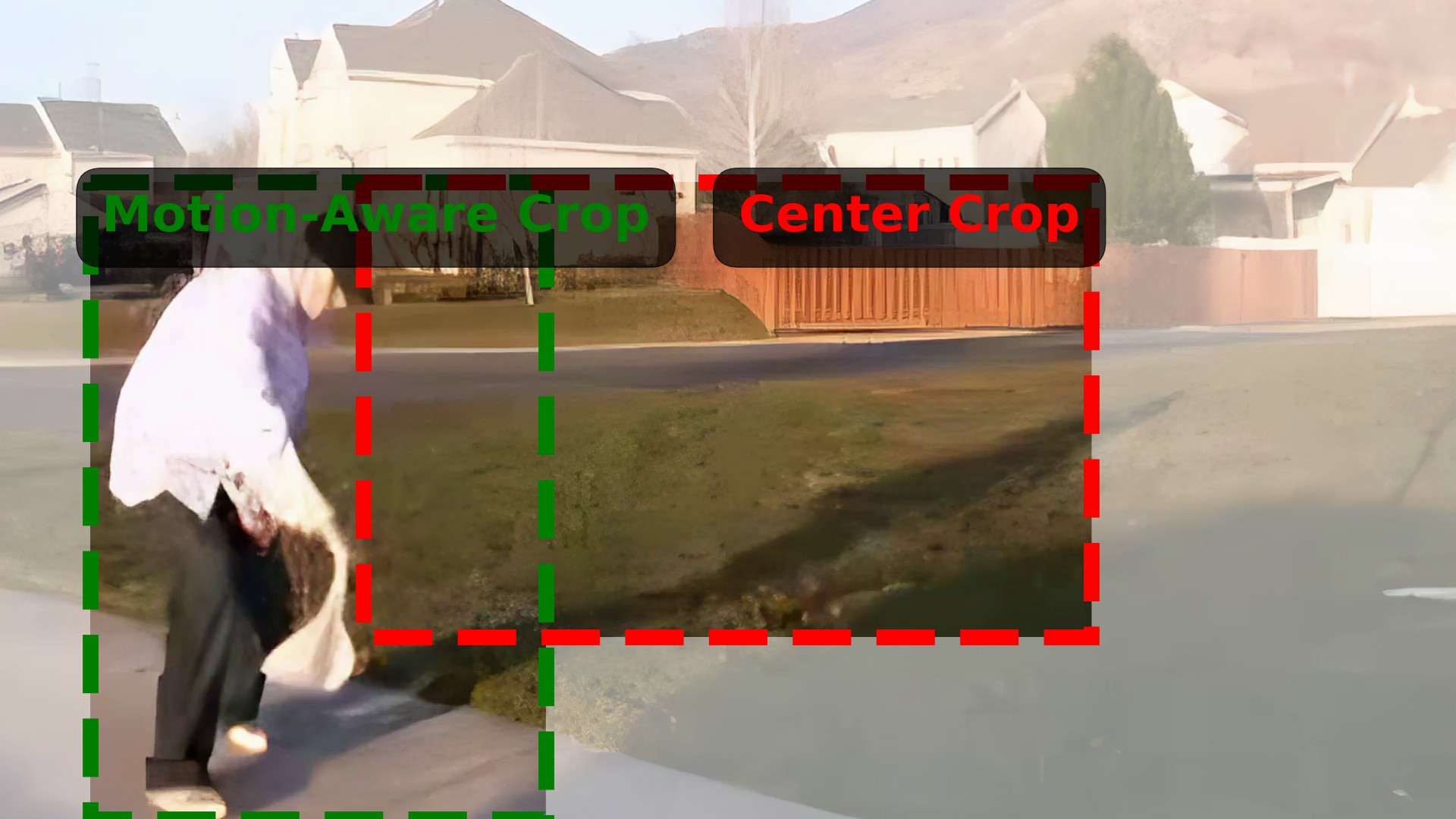} \\
\multicolumn{6}{l}{\textit{PizzaTossing}} \\[4pt]

\includegraphics[width=0.15\textwidth]{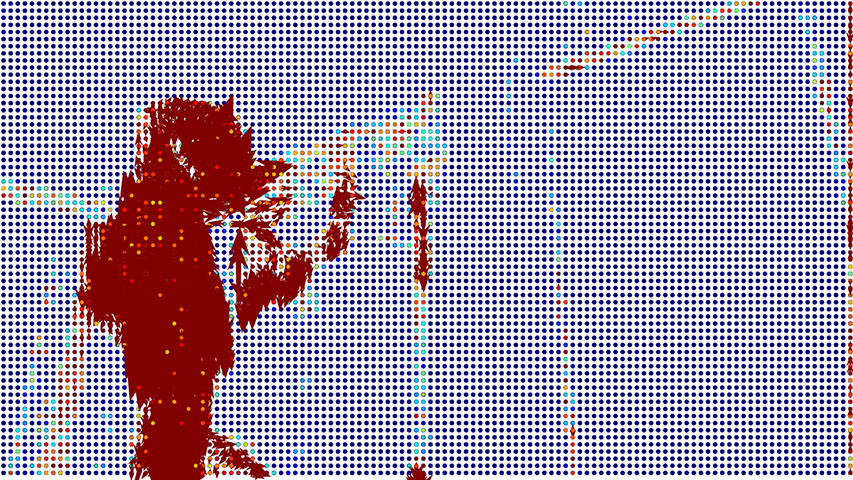} &
\includegraphics[width=0.15\textwidth]{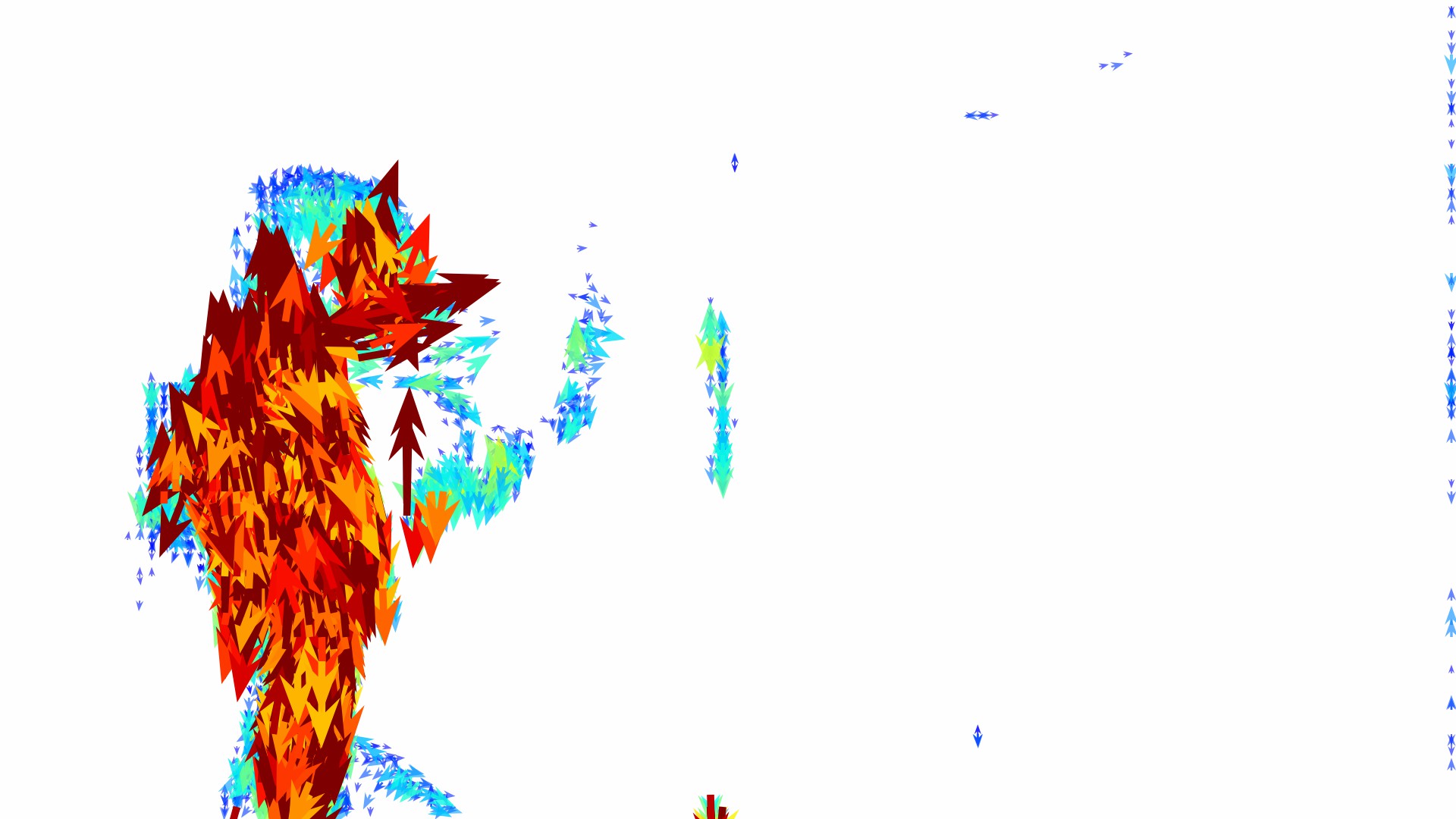} &
\includegraphics[width=0.15\textwidth]{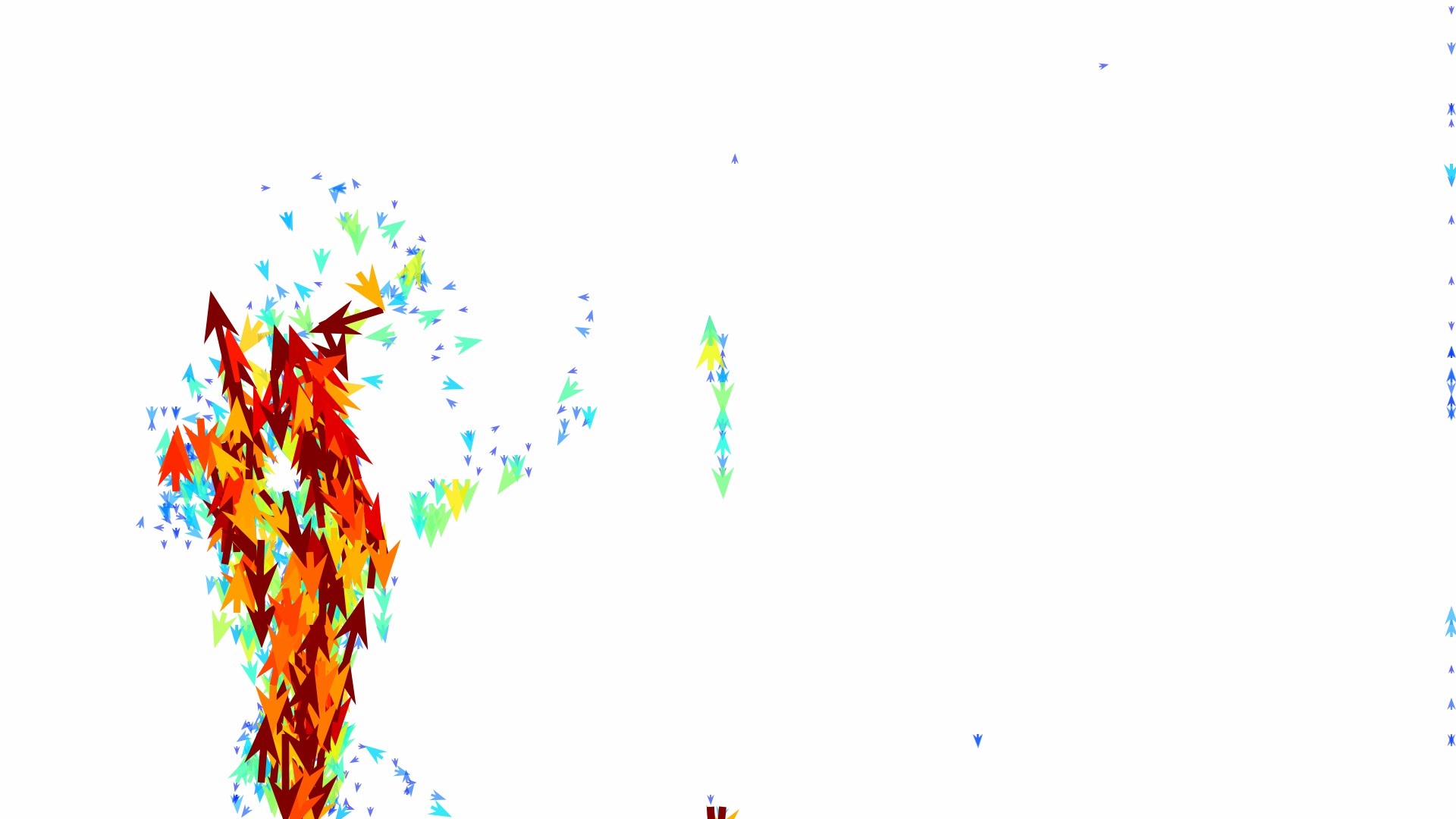} &
\includegraphics[width=0.15\textwidth]{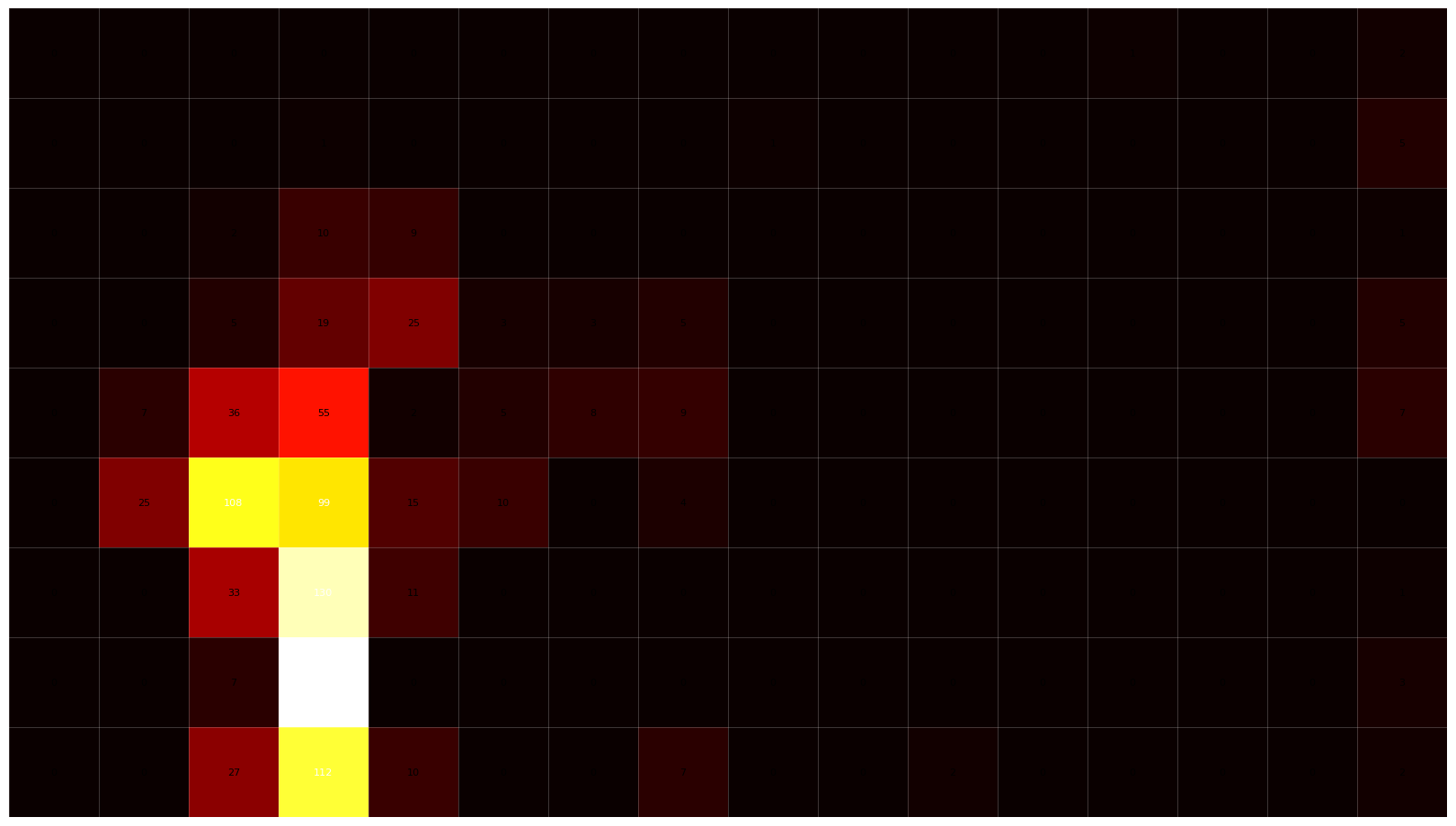} &
\includegraphics[width=0.15\textwidth]{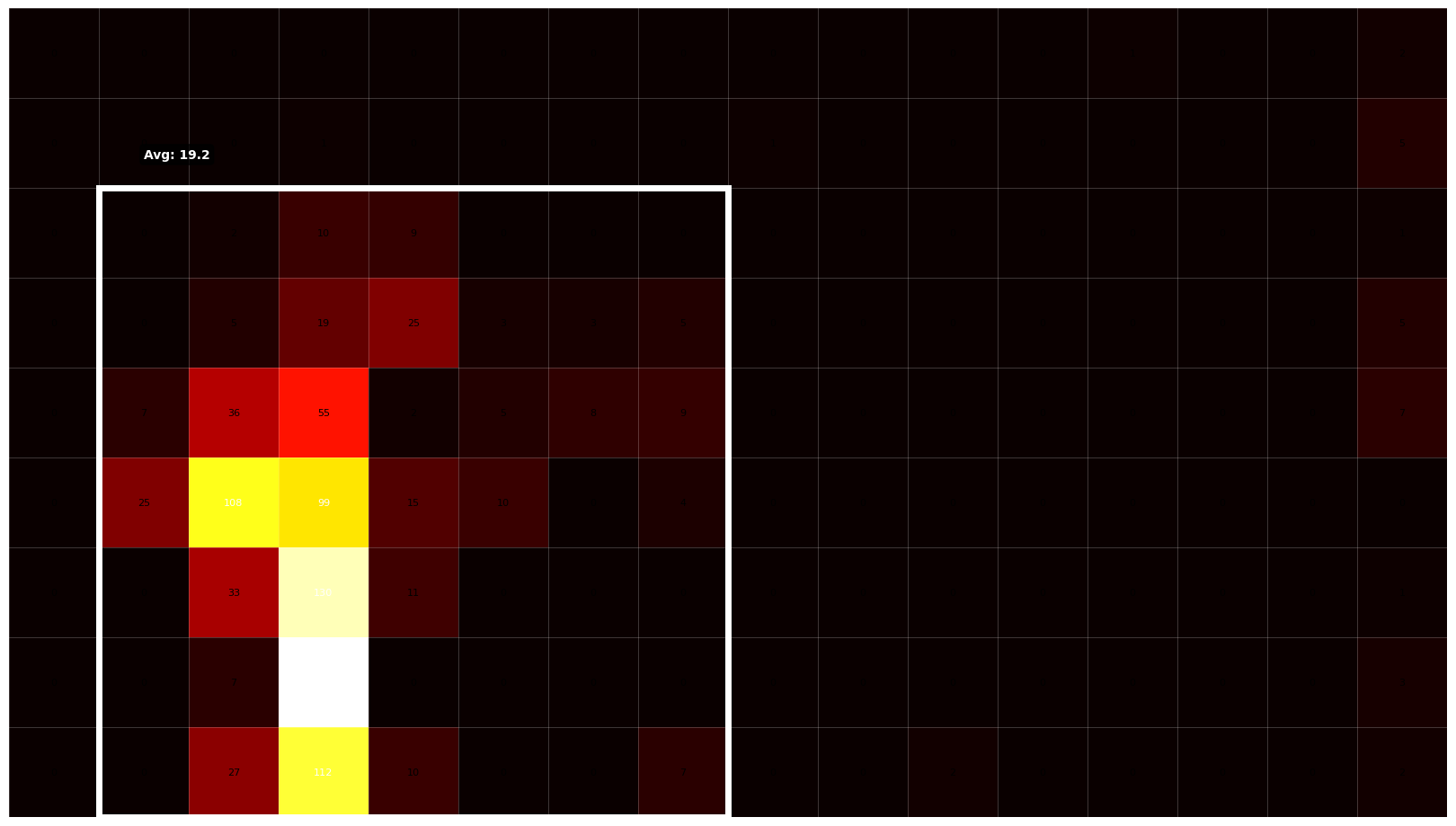} &
\includegraphics[width=0.15\textwidth]{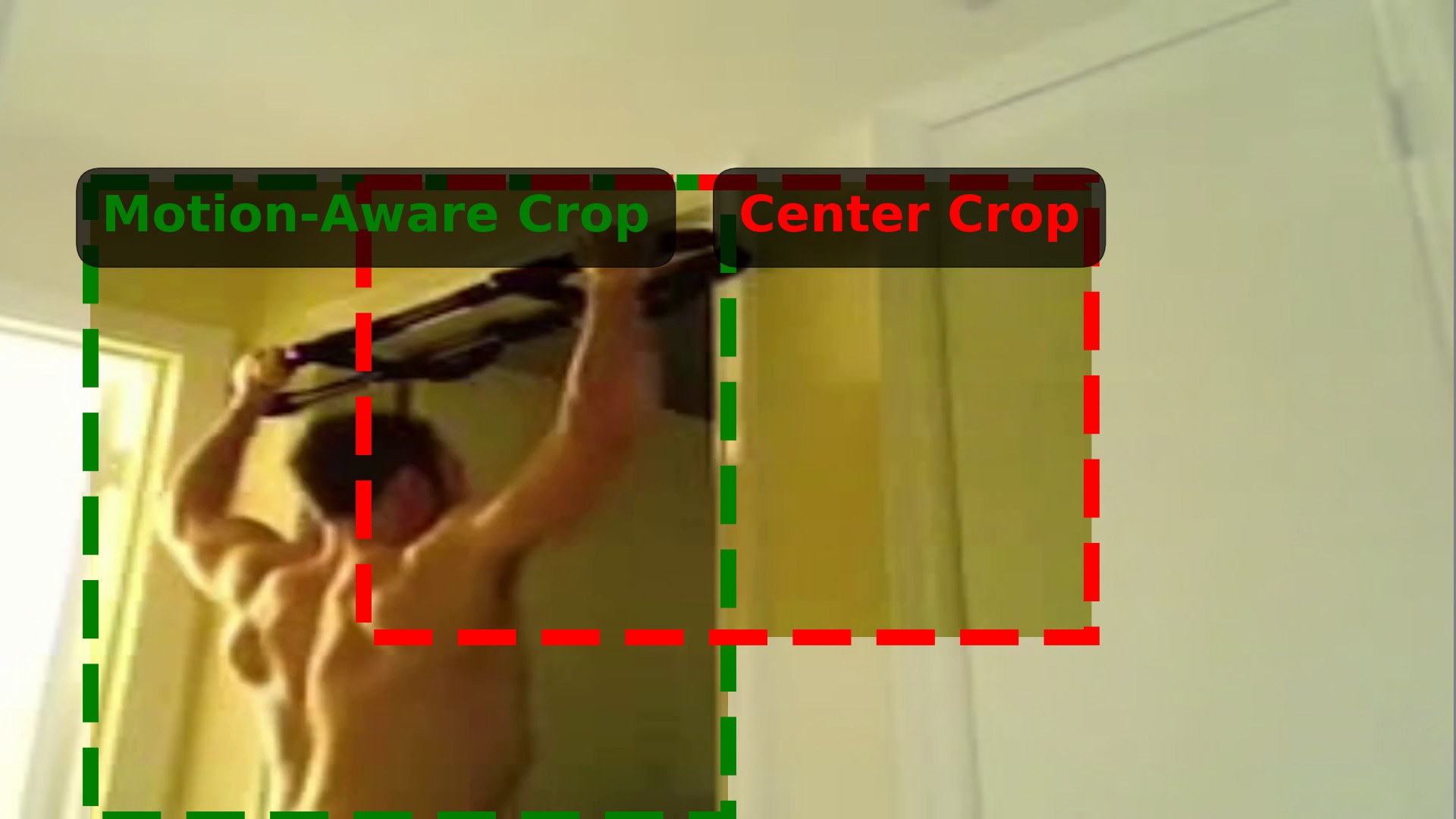} \\
\multicolumn{6}{l}{\textit{PullUps}} \\[4pt]

\end{tabular}

\caption{Qualitative visualization of the complete MoCrop pipeline across six representative UCF101 action classes. Each row shows one video processed through six stages: 
\textbf{(a) Raw MVs}: MVs extracted from H.264 compressed video, visualized as arrows showing magnitude and direction. 
\textbf{(b) MD (Denoised)}: Top-1\% MVs retained after percentile-based filtering. 
\textbf{(c) MCS (Sampled)}: MVs after weighted importance sampling (10\% of denoised MVs, biased toward high-magnitude motion). 
\textbf{(d) MGS - Grid}: Motion-density heatmap aggregated on $16{\times}9$ spatial grid. 
\textbf{(e) MGS - Search}: Optimal region identified by dual-objective scoring (white box overlaid on heatmap). 
\textbf{(f) Comparison}: Overlay comparison of Motion-Aware Crop (green box) vs. Center Crop (red box) on the same RGB frame, demonstrating MoCrop's ability to focus on action-relevant regions.}
\label{fig:pipeline_complete}
\end{figure*}

\begin{figure}[t]
\centering
\scriptsize
\setlength{\tabcolsep}{1pt}
\begin{tabular}{ccc}
\textbf{(a) Motion-density} & \textbf{(b) Overlay} & \textbf{(c) Comparison} \\[2pt]
\includegraphics[width=0.33\columnwidth, height=1.8cm, keepaspectratio=false]{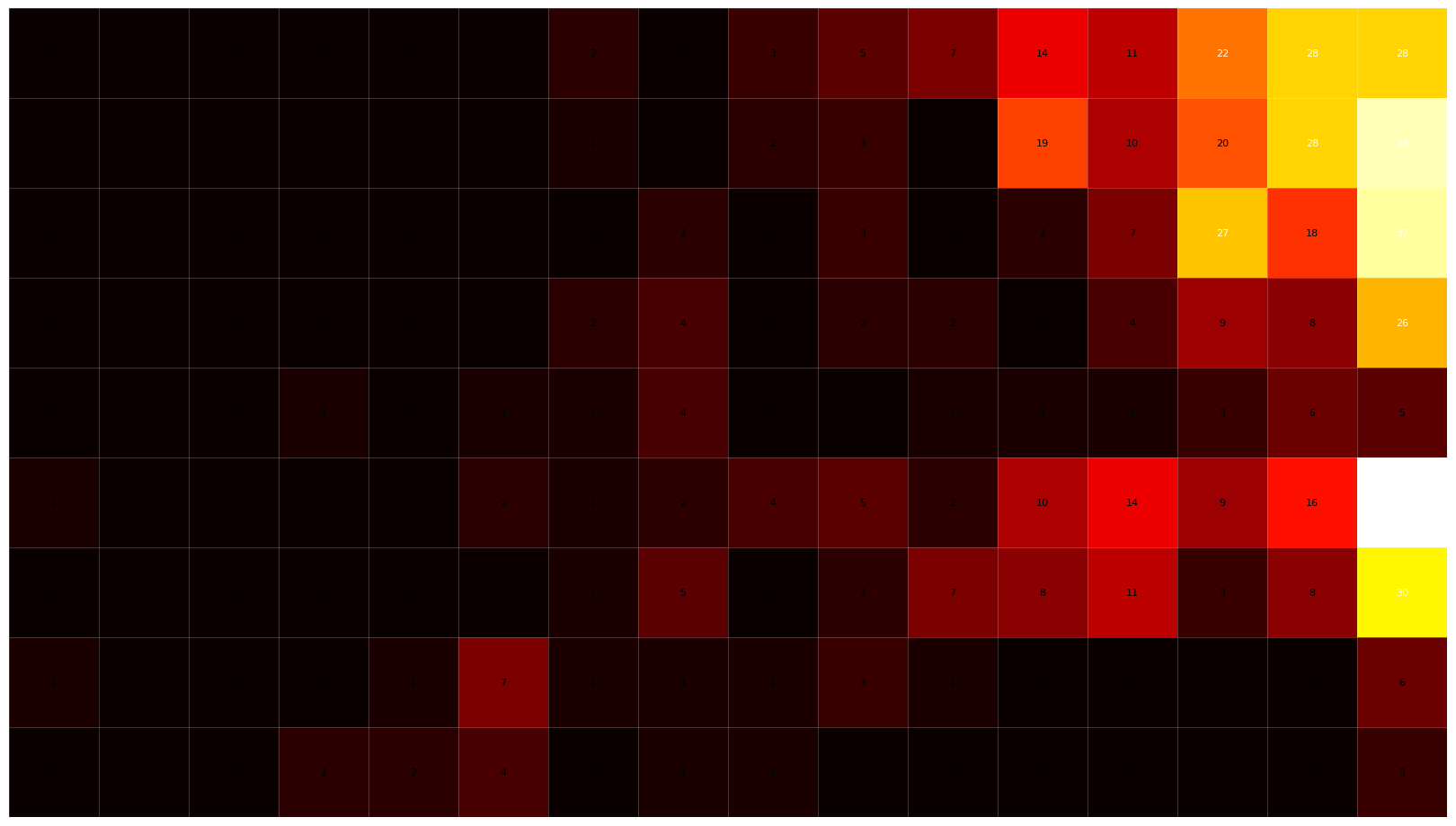} &
\includegraphics[width=0.33\columnwidth, height=1.8cm, keepaspectratio=false]{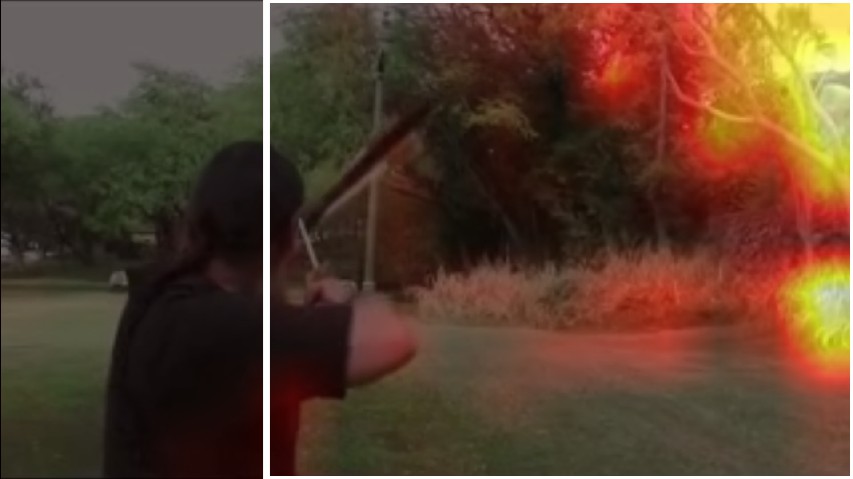} &
\includegraphics[width=0.33\columnwidth, height=1.8cm, keepaspectratio=false]{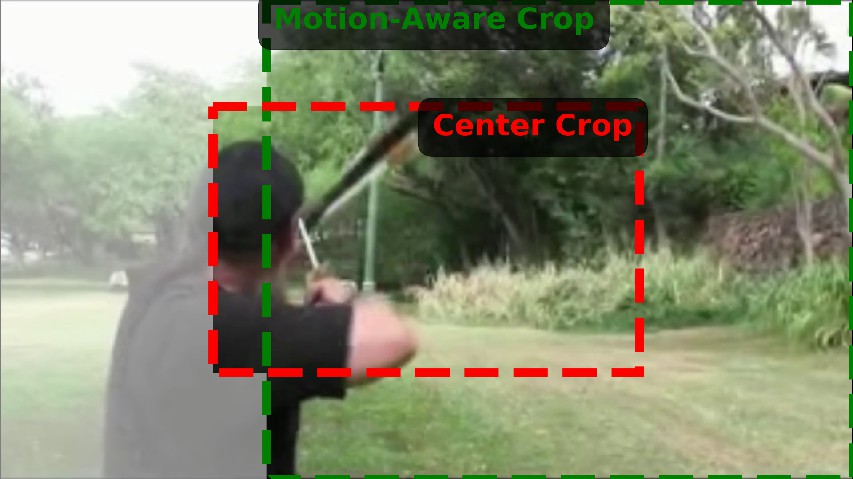} \\
\multicolumn{3}{l}{\parbox{\columnwidth}{\textit{Archery: Bright-corner artifact from camera shake.}}} \\[4pt]
\includegraphics[width=0.33\columnwidth, height=1.8cm, keepaspectratio=false]{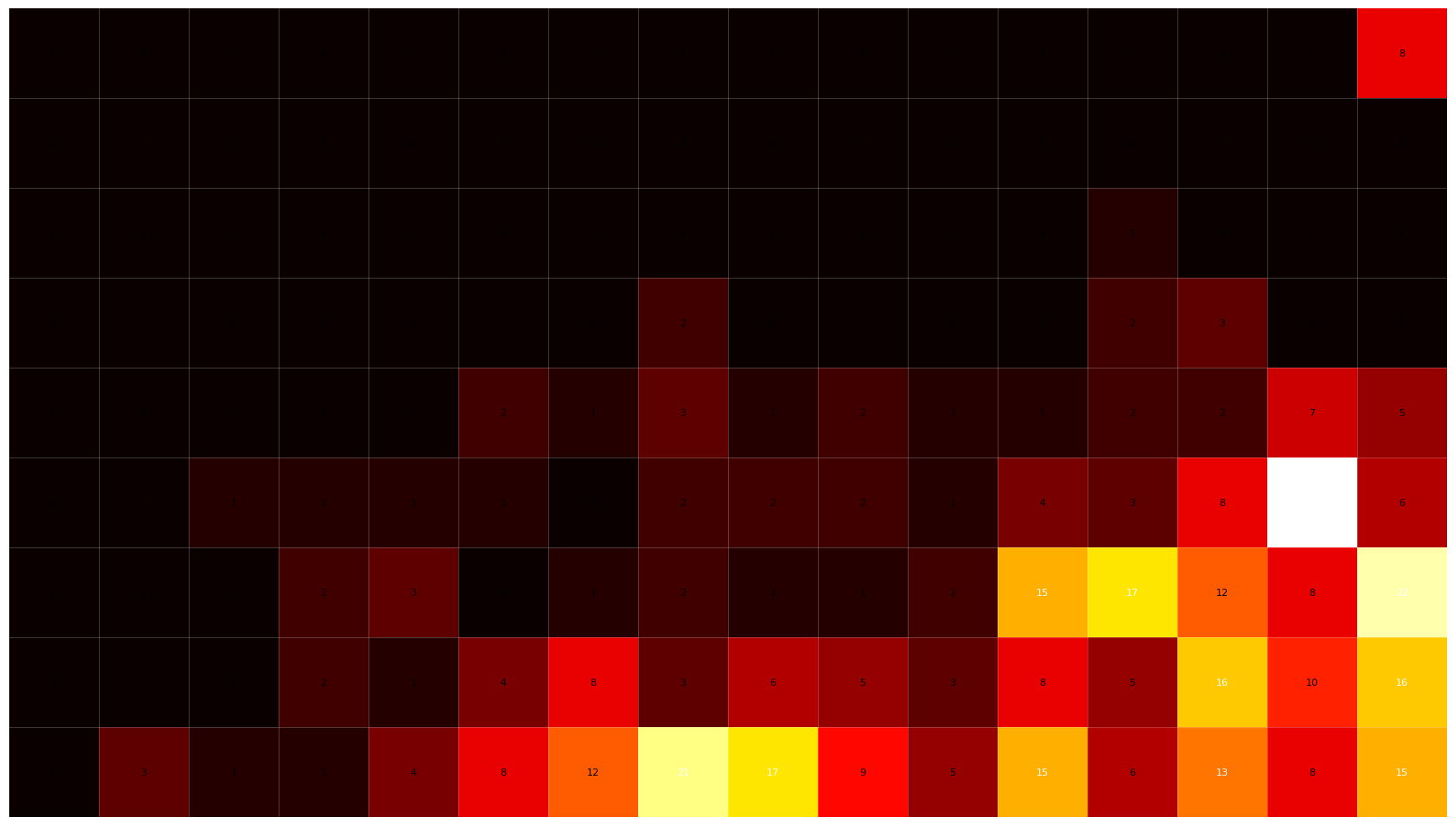} &
\includegraphics[width=0.33\columnwidth, height=1.8cm, keepaspectratio=false]{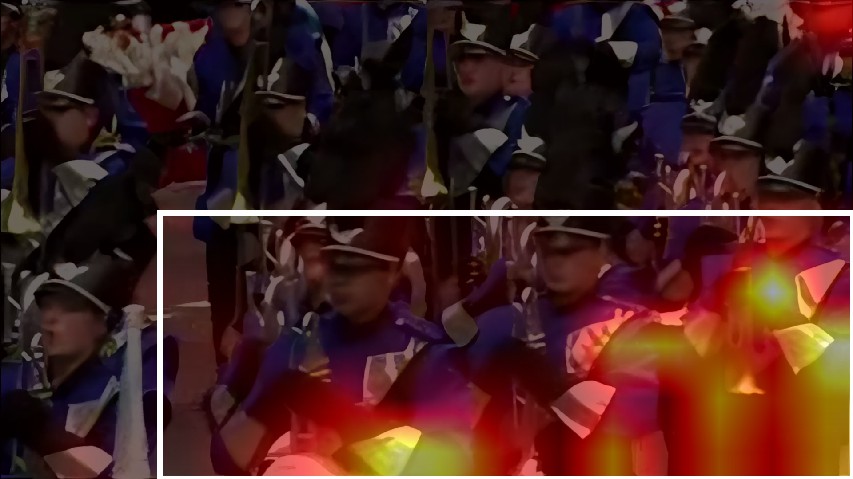} &
\includegraphics[width=0.33\columnwidth, height=1.8cm, keepaspectratio=false]{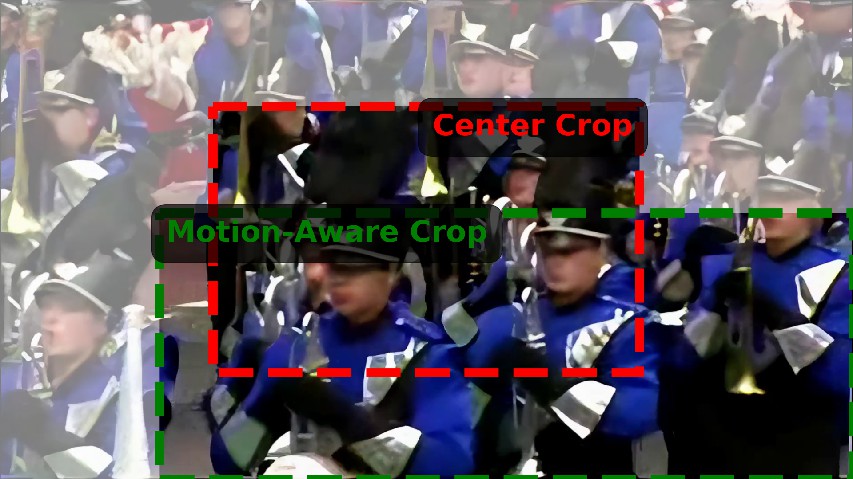} \\
\multicolumn{3}{l}{\parbox{\columnwidth}{\textit{BandMarching: Median rescaling biases toward near-camera region.}}} \\[4pt]
\includegraphics[width=0.33\columnwidth, height=1.8cm, keepaspectratio=false]{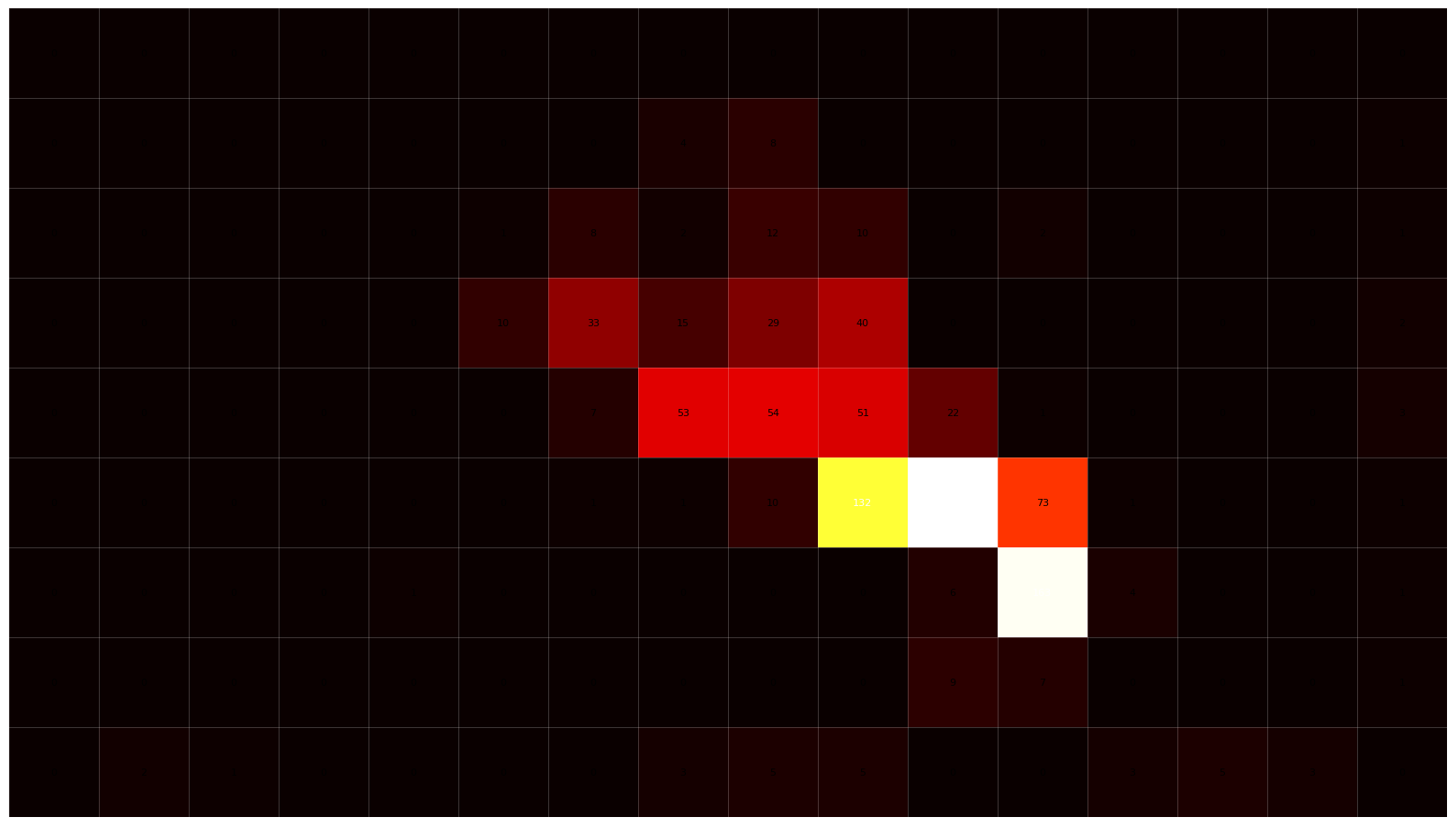} &
\includegraphics[width=0.33\columnwidth, height=1.8cm, keepaspectratio=false]{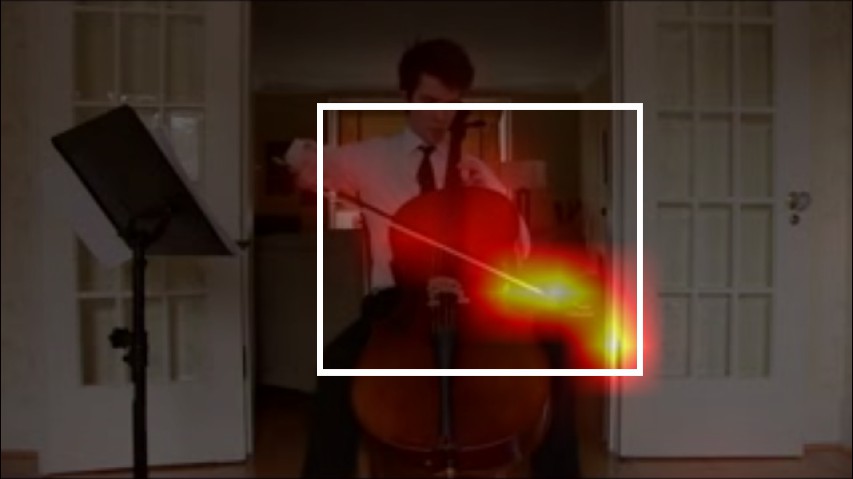} &
\includegraphics[width=0.33\columnwidth, height=1.8cm, keepaspectratio=false]{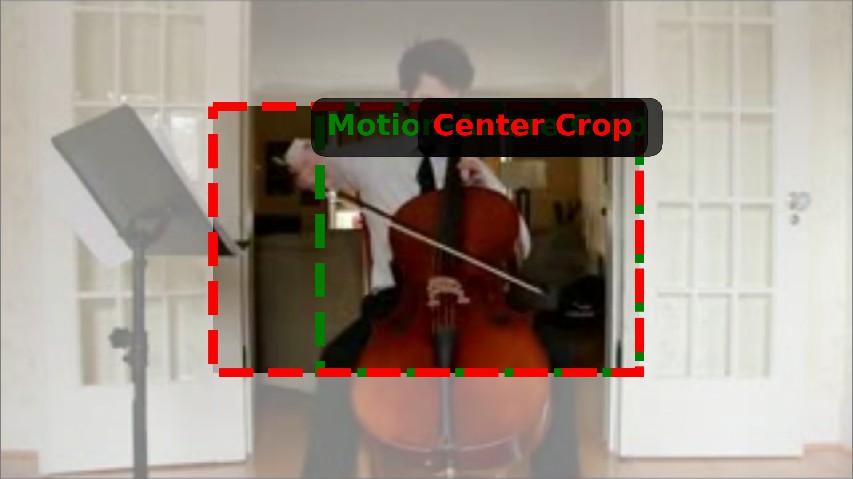} \\
\multicolumn{3}{l}{\parbox{\columnwidth}{\textit{PlayingCello: Centered actor; only bow motion detected.}}} \\[4pt]
\includegraphics[width=0.33\columnwidth, height=1.8cm, keepaspectratio=false]{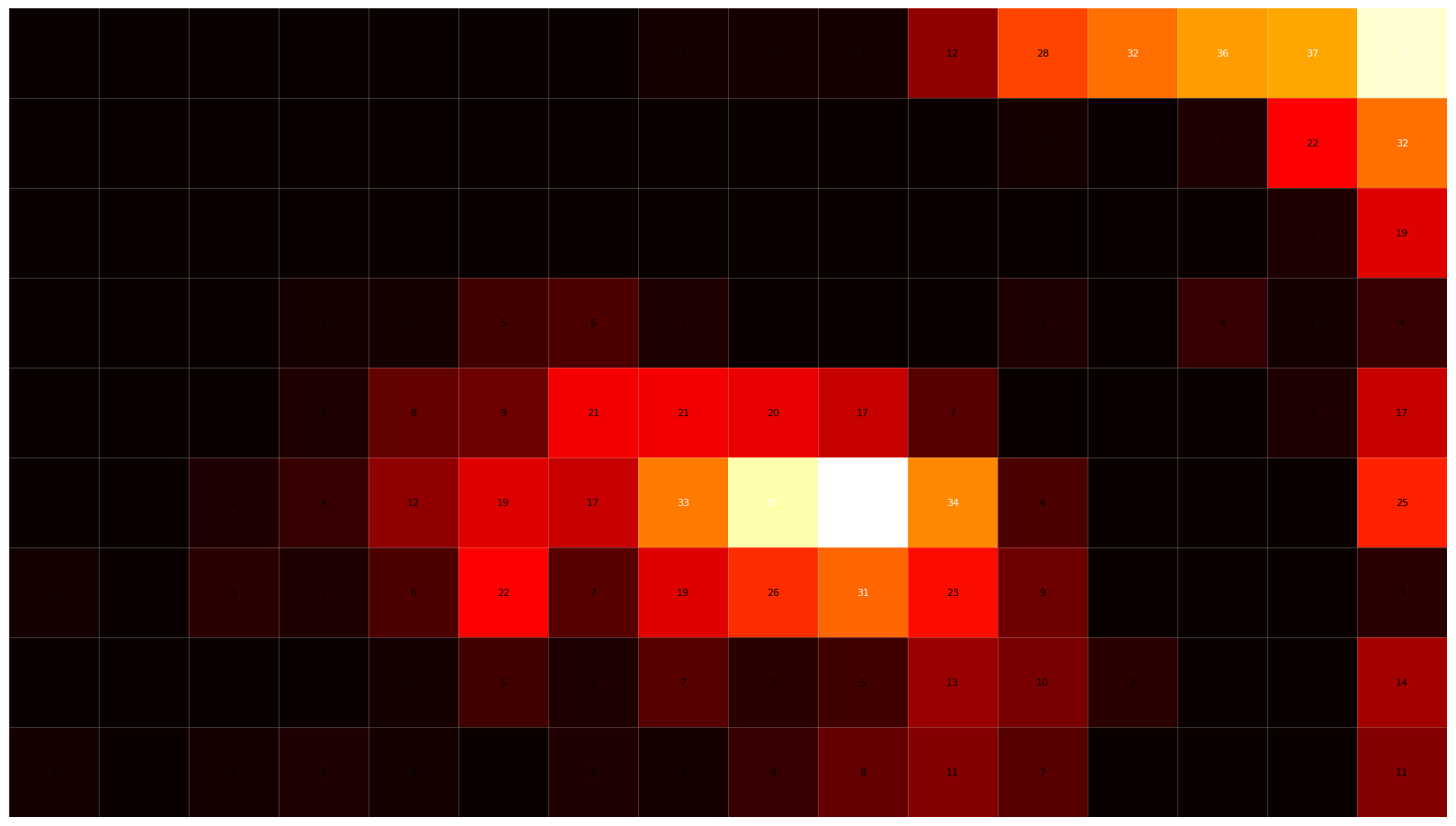} &
\includegraphics[width=0.33\columnwidth, height=1.8cm, keepaspectratio=false]{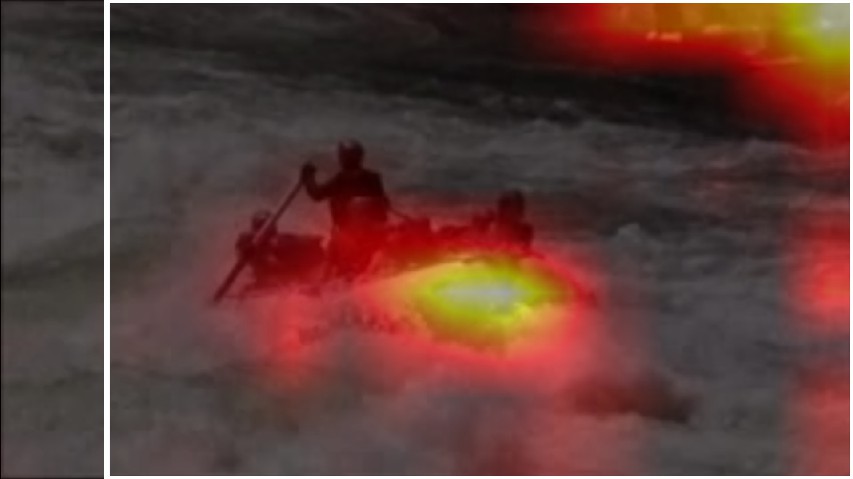} &
\includegraphics[width=0.33\columnwidth, height=1.8cm, keepaspectratio=false]{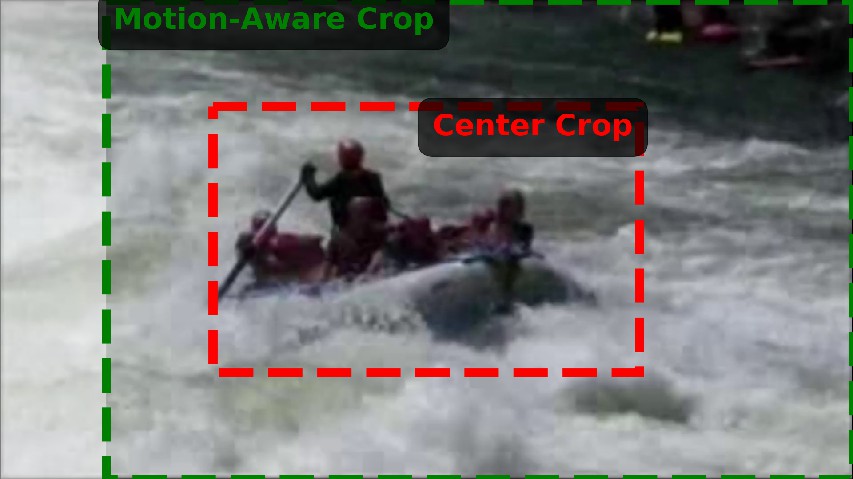} \\
\multicolumn{3}{l}{\parbox{\columnwidth}{\textit{Rafting: Water motion fills the entire frame.}}} \\[4pt]
\end{tabular}
\caption{Failure mode analysis across five representative cases. (a) Motion-density map. (b) Heatmap overlay with search result (white box). (c) Motion-Aware Crop (green) vs. Center Crop (red). See text for detailed analysis.}
\label{fig:failure_cases}
\end{figure}

\section{Discussion and Conclusion}
\label{sec:discussion_conclusion}

We present MoCrop, a lightweight, training-free module using compressed-domain MVs to identify motion-dense regions for adaptive cropping. By focusing inference, MoCrop reduces resolution and FLOPs, often boosting accuracy, and integrates with existing backbones without retraining.

\noindent\textbf{When it helps.}
MoCrop is most effective when actions coincide with localized motion and when background clutter dilutes fixed crops. The map-based formulation is robust to moderate noise after filtering and sampling, and it generalizes across backbones since it operates as a preprocessing step.

\noindent\textbf{Limitations.}
Fig.~\ref{fig:failure_cases} illustrates five failure modes where MoCrop's motion-density approach encounters challenges. 
MoCrop relies on codec-provided MVs (e.g., H.264) and is less informative for raw videos unless motion estimation is run in parallel. This dependency is acceptable since most datasets and streaming pipelines use compressed video. 

Specifically, the following scenarios can bias the density map:
\textbf{(1) Camera shake} (\textit{Archery}): When the cameraman shakes the lens, spurious high-magnitude MVs appear (e.g., bright regions in corners), dominating the density map and drawing the crop away from the actual actor.
\textbf{(2) Perspective effects} (\textit{BandMarching}): When subjects move uniformly across the frame, perspective makes closer objects move faster. Since our pipeline rescales MVs by clip-wise median magnitude for robustness, this causes the near-camera region to dominate, yielding incorrect crops.
\textbf{(3) Stationary actors} (\textit{PlayingCello}): When actors are already centered with minimal body translation, only small regions (e.g., the bow) exhibit motion. The density map localizes these regions but fails to cover the full actor, offering little advantage over center cropping.
\textbf{(4) Scene-wide motion} (\textit{Rafting}): Turbulent water or dynamic backgrounds generate MVs across the entire frame, producing nearly uniform density maps. MGS then selects regions close to the full frame, providing no effective spatial focus.

Mitigations include global-motion suppression, temporal smoothing, and combining MV magnitude with residual cues. These failure modes occur in a minority of UCF101 action classes and primarily arise when motion does not reliably indicate actor location.

\noindent\textbf{Future work.}
We plan to extend MoCrop to diagonally elongated or multi-region actions via multi-box proposals, explore codec-agnostic motion signals, and validate on larger datasets and backbones. We will also study learned selection over MoCrop proposals while keeping inference cost low. 

In summary, MoCrop offers a practical plug-and-play path to efficiency and accuracy in compressed-domain video understanding, requiring no additional parameters and minimal compute overhead.

\bibliographystyle{IEEEbib}
\bibliography{strings,refs}

\end{document}